\documentclass[10pt,twocolumn,letterpaper]{article}

\usepackage{cvpr}      

\usepackage[dvipsnames]{xcolor}

\definecolor{cvprblue}{rgb}{0.21,0.49,0.74}
\usepackage[pagebackref,breaklinks,colorlinks,citecolor=cvprblue]{hyperref}

\usepackage{natbib}
\setcitestyle{numbers,square}

\usepackage{algorithm}
\usepackage{algorithmic}

\usepackage{bm}
\usepackage{booktabs}
\usepackage{threeparttable}
\usepackage{multirow}

\usepackage{amsmath}
\usepackage{amssymb}
\usepackage{tabu}

\title{ComFusion: Personalized Subject Generation in Multiple Specific Scenes From Single Image}
\author{\textnormal{Yan Hong}$^{1}$, \textnormal{Yuxuan Duan}$^{2}$, \textnormal{Bo Zhang}$^{2}$, \textnormal{Haoxing Chen}$^{1}$, \textnormal{Jun Lan}$^{1}$, \\  \textnormal{Huijia Zhu}$^{1}$, \textnormal{Weiqiang Wang}$^{1}$, \textnormal{Jianfu Zhang}$^{3}$\thanks{Corresponding author.}~\\
$^{1}$Ant Group, 
$^{2}$MoE Key Lab of Artificial Intelligence, Shanghai Jiao Tong University \\
$^{3}$Qing Yuan Research Institute, Shanghai Jiao Tong University \\
$^{1}${\tt\small yanhong.sjtu@gmail.com, hx.chen@hotmail.com}, $^{2}${\tt\small \{sjtudyx2016,bo-zhang\}@sjtu.edu.cn}  \\
$^{1}${\tt\small \{yelan.lj, huijia.zhj, weiqiang.wwq\}@antgroup.com}, $^{3}${\tt\small c.sis@sjtu.edu.cn}
}

\begin{document}
\maketitle
\begin{abstract}
Recent advancements in personalizing text-to-image (T2I) diffusion models have shown the capability to generate images based on personalized visual concepts using a limited number of user-provided examples.
However, these models often struggle with maintaining high visual fidelity, particularly in manipulating scenes as defined by textual inputs.
Addressing this, we introduce ComFusion, a novel approach that leverages pretrained models generating composition of a few user-provided subject images and predefined-text scenes, effectively fusing visual-subject instances with textual-specific scenes, resulting in the generation of high-fidelity instances within diverse scenes.
ComFusion integrates a class-scene prior preservation regularization, which leverages composites the subject class and scene-specific knowledge from pretrained models to enhance generation fidelity. 
Additionally, ComFusion uses coarse generated images, ensuring they align effectively with both the instance image and scene texts. 
Consequently, ComFusion maintains a delicate balance between capturing the essence of the subject and maintaining scene fidelity.
Extensive evaluations of ComFusion against various baselines in T2I personalization have demonstrated its qualitative and quantitative superiority.
\end{abstract}

\begin{figure}[htp]
\begin{center}
\includegraphics[width=0.9\linewidth]{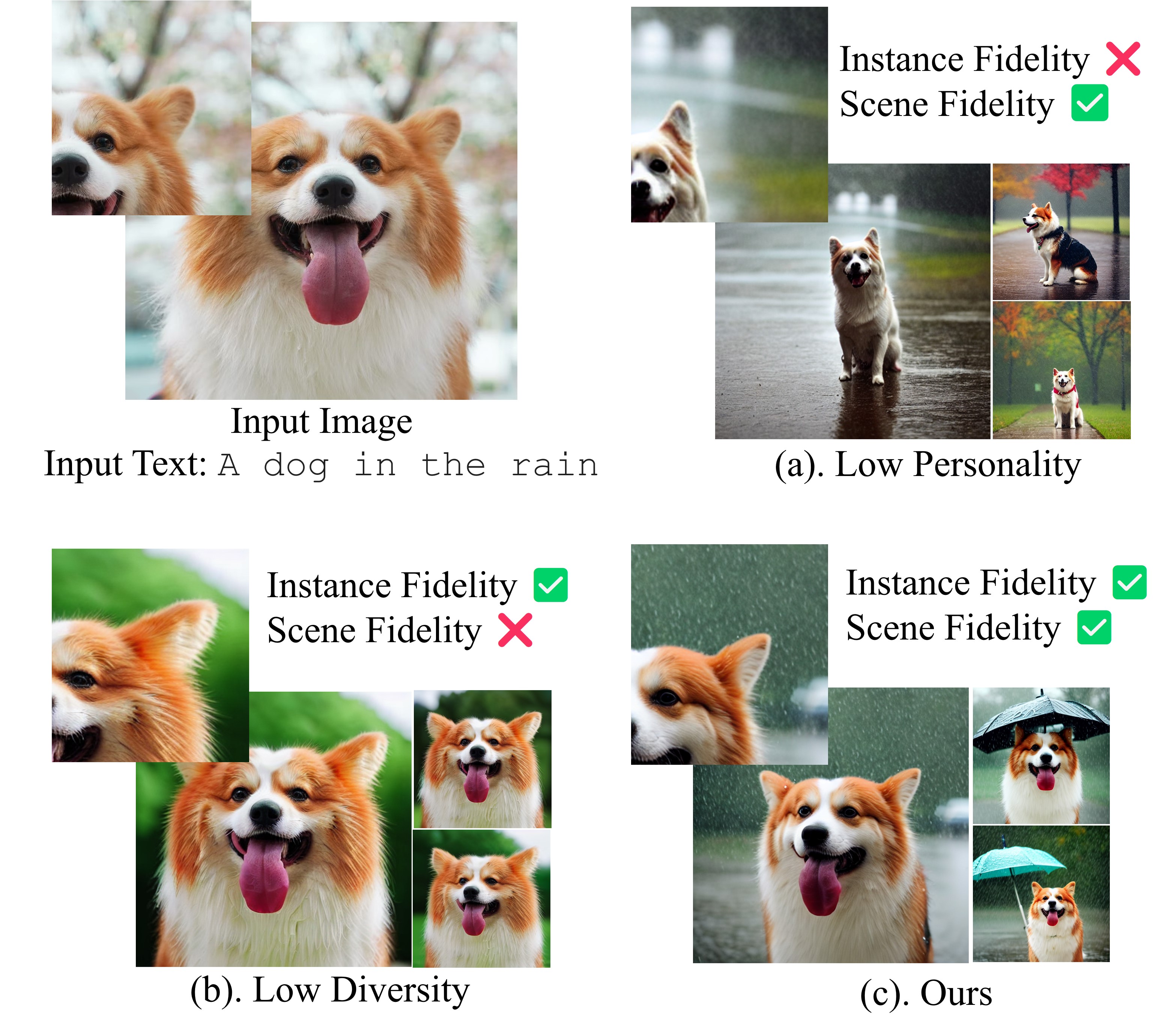}
\end{center}
\vspace{-16pt}
\caption{
Contrasting with existing methods \cite{ruiz2023dreambooth,avrahami2023break} , which often face challenges in simultaneously preserving instance fidelity and scene fidelity, ComFusion skillfully composites the instance image with textual prompts and fuses the visual details of the subject instance with the textual variations of the scenes, yielding the creation of plausible, personalized images that exhibit a rich diversity.
}
\vspace{-8pt}
\label{fig:motivation} 
\end{figure}

\section{Introduction}
Text-to-image (T2I) personalization aims to customize a diffusion-based T2I model with user-provided visual concepts~\cite{ruiz2023dreambooth,gal2022image}.
This innovative approach enables the creation of new images that seamlessly integrate these concepts into diverse scenes. 
More formally, given a few images of a subject (no more than five), our objective is embed this subject into the model's output domain. This allows for the synthesis of the subject with a unique identifier in various scenes. 
The task of rendering such imaginative scenes is particularly challenging. It entails the synthesis of specific subjects (\textit{e.g.}, objects, animals) in new contexts,  ensuring their natural and seamless incorporation into the scene. Such a task demands a delicate balance between the subject's distinctive features and the new scene context.

Recently, this field of T2I personalization has attracted significant attention from the academic community with many works \cite{avrahami2023break,kumari2023multi,gal2022image,voynov2023p+,wei2023elite,tewel2023key,arar2023domain,
shi2023instantbooth} leveraging the capabilities of advanced diffusion-based T2I models \cite{rombach2022high,ramesh2022hierarchical,saharia2022photorealistic}. 
These approaches broadly fall into two categories:
The first category \cite{gal2023encoder,chen2023subject,shi2023instantbooth,li2023blip,jia2023taming} integrates additional modules with a pretrained base model. This stream enables the creation of personalized subjects without the need for finetuning during testing. However, it often struggles to maintain the subject's identity consistently across different synthesized images.
In contrast, the second category \cite{ruiz2023dreambooth,gal2022image,kumari2023multi,avrahami2023break} infocuses on finetuning the pretrained model with a select set of images, employing various regularization techniques and training strategies. 
This finetuning process effectively harnesses the model’s existing class knowledge, combined with the unique identifier of the subject, thereby allowing for the generation of diverse variations of the subject in various contexts.

However, finetuning diffusion models presents a challenge known as \textit{language drift} \cite{lu2020countering,lee2019countering}, where the finetuned model may lose its pre-finetuning acquired knowledge, including the understanding of a diverse range of classes and scenes.
Furthermore, few-shot learning paradigms are susceptible to \textit{catastrophic neglecting}, particularly when generating new images with specific scenes, leading to inadequate generation or integration of some prompts or subjects. 
These issues contribute to a notable decrease in both \textit{scene fidelity} and \textit{instance fidelity}.
In ~\cref{fig:motivation}, we show an example of the personalized generation using a specific dog instance image and the text prompt ``\texttt{A dog in the rain}''. The images, generated by existing leading methods \cite{ruiz2023dreambooth,avrahami2023break} and our proposed approach.
 \cref{fig:motivation} (a) illustrates a lack of \textit{instance fidelity}, where the generated images fail to preserve the subject dog's appearance, resulting in low-personality output.
 \cref{fig:motivation} (b) highlights examples with insufficient \textit{scene fidelity}, failing to accurately represent the rainy scene, thus limiting the diversity of the generated images.

To address the issues of language drift and catastrophic neglecting and to improve the instance fidelity and scene fidelity of generated images, we propose \textbf{ComFusion} (\textbf{Com}posite and \textbf{Fusion}). 
This innovative approach is tailored to enable personalized subject generation across diverse scenes. 
ComFusion utilizes a finetuning strategy to effectively composite and fuse visual subject features with textual features, thus synthesizing new images with high-fidelity instances composed with a variety of distinct scenes.
To achieve this, we have established two streams: a \textit{composite stream} and a \textit{fusion stream}. 
Within the composite stream, we introduce \textit{class-scene prior loss} to preserve the pretrained model’s knowledge of class and scene.
This strategy leads to the generation of a diverse array of images that \textit{composite} the essence of class and scene priors with subject instances and their contexts, alleviating language drift in scene representation, enhancing coherent syntheses of both subject instances and scene contexts.
In the fusion stream, we propose a \textit{visual-textual matching loss} to effectively \textit{fuse} the visual information of the subject instance with the textual information of the scene, ensuring their integrated representation in the coarsely generated images.
Through the fusion stream, ComFusion mitigates the catastrophic neglecting issue, achieving a harmonious balance between instance fidelity and scene fidelity.
In  \cref{fig:motivation}, we present some impressive samples obtained by ComFusion. The images illustrate a remarkable preservation of the dog's appearance, while the scene ‘\texttt{in the rain}’ is brought to life with vivid details such as rain spots and umbrellas. 
To further substantiate ComFusion's effectiveness, we have conducted extensive experiments across various subject instances and scenes. These studies confirm that ComFusion excels both qualitatively and quantitatively, outperforming existing methods.

\section{Related Works}
\noindent\textbf{Diffuion-Based Text-to-Image Generation}:
The field of Text-to-Image (T2I) generation has recently witnessed remarkable advancements ~\cite{zhang2023adding,kang2023scaling,tao2022df,tao2023galip,saharia2022photorealistic,gu2022vector}, predominantly led by pre-trained diffusion models such as Stable Diffusion~\cite{rombach2022high}, DALLE~\cite{ramesh2022hierarchical}, Imagen~\cite{saharia2022photorealistic} and etc.
These models are renowned for their exceptional control in producing photorealistic images that closely align with textual descriptions.
Despite their superior capabilities in generating high-quality images, these models encounter challenges in more personalized image generation tasks 
, which are often difficult to be precisely described with text descriptions. 
This challenge has sparked interest in the rapidly evolving field of personalized T2I generation~\cite{ruiz2023dreambooth,gal2022image, kumari2023multi,tewel2023key,liu2023cones}.

\noindent\textbf{Personalized Text-to-Image Generation}:
Given a small set of images of the subject concept, personalized T2I generation~\cite{ruiz2023dreambooth,gal2022image,kumari2023multi,gal2023encoder,voynov2023p+,liu2023cones,wei2023elite,tewel2023key,shi2023instantbooth,arar2023domain,chen2023disenbooth,ho2020denoising,song2020denoising,ramesh2022hierarchical,saharia2022photorealistic} aims to generate new images according to the text descriptions while maintaining the identity of the subject concept. 
Early studies in training generative models in few-shot setting focus on alleviating mode collapse~\cite{srivastava2017veegan,liu2019spectral,thanh2020catastrophic} for generative adversarial networks~\cite{clouatre2019figr,hong2020matchinggan, ding2022attribute,yang2022wavegan,hong2020f2gan, hong2020deltagan,li2023euclidean}. 
Recently, finetuning diffusion-based text-to-image model with a few images has also been explored in ~\cite{ruiz2023dreambooth,avrahami2023break}. 
In stream of diffusion-based generators, personalized T2I generation methods can be classified into two categories: 
The first stream involves the integration of additional modules (\textit{e.g.}, ~\cite{mou2023t2i,hu2021lora,zhang2023adding}) with a pretrained base model. 
The second stream adopts a strategy of finetuning the pretrained model using a few selected images.

\noindent{\textbf{Without Finetuning}}: 
These methods ~\cite{gal2023encoder,chen2023subject,shi2023instantbooth,li2023blip,jia2023taming} generally rely on additional modules trained on additional new datasets, such as the visual encoder in~\cite{shi2023instantbooth,wei2023elite} and the experts in~\cite{chen2023subject,li2023blip} to directly map the image of the new subject to the textual space.  
Specifically, \cite{gal2023encoder} introduces an encoder that encodes distinctive instance information, enabling rapid integration of novel concepts from a given domain by training on a diverse range of concepts within that domain.
In \cite{shi2023instantbooth}, a learnable image encoder translates input images into textual tokens, supplemented by adapter layers in the pre-trained model, thus facilitating rich visual feature representation and instant text-guided image personalization without requiring test-time finetuning. 
DisenBooth \cite{chen2023disenbooth} uses weak denoising and contrastive embedding auxiliary tuning objectives for personalization.  
ELITE~\cite{wei2023elite}  introduces a method for learning both local and global maps on large-scale datasets, allowing for instant adaptation to unseen instances using a single image marked with the subject concept for personalized generation. 

\noindent{\textbf{Finetuning}}: 
Various methods employ diverse training strategies to optimize different modules within pretrained models ~\cite{ruiz2023dreambooth,gal2022image,kumari2023multi}.
DreamBooth~\cite{ruiz2023dreambooth} and TI~\cite{gal2022image} are two popular subject-driven text-to-image generation methods based on finetuning. Both approaches map subject images to a special prompt token during finetuning. They differ in their finetuning focus: TI concentrates on prompt embedding, while DreamBooth targets the U-Net model and text-encoder.
Recent finetuning-based methods~\cite{kumari2023multi,voynov2023p+} focus on how to design training strategy to update core parameters of T2I model for subject concepts on user-provided 4-6 images. 
A domain-agnostic method is proposed in \cite{arar2023domain} that introduce a novel contrastive-based regularization technique. This technique aims to preserve high fidelity to the subject concept's characteristics while keeping the predicted embeddings close to editable regions of the latent space. 
Break-A-Scene \cite{avrahami2023break} utilizes the subject concept's mask and employs a two-stage process for personalized T2I generation using a single image. However, this approach is limited in terms of the subject's diversity.
In this paper, we concentrate on advancing personalized T2I generation. Our goal is to establish an ideal balance between fidelity to the subject concept and adaptability to multiple specific scenes.

\begin{figure*}[htp]
\begin{center}
\includegraphics[width=0.9\linewidth]{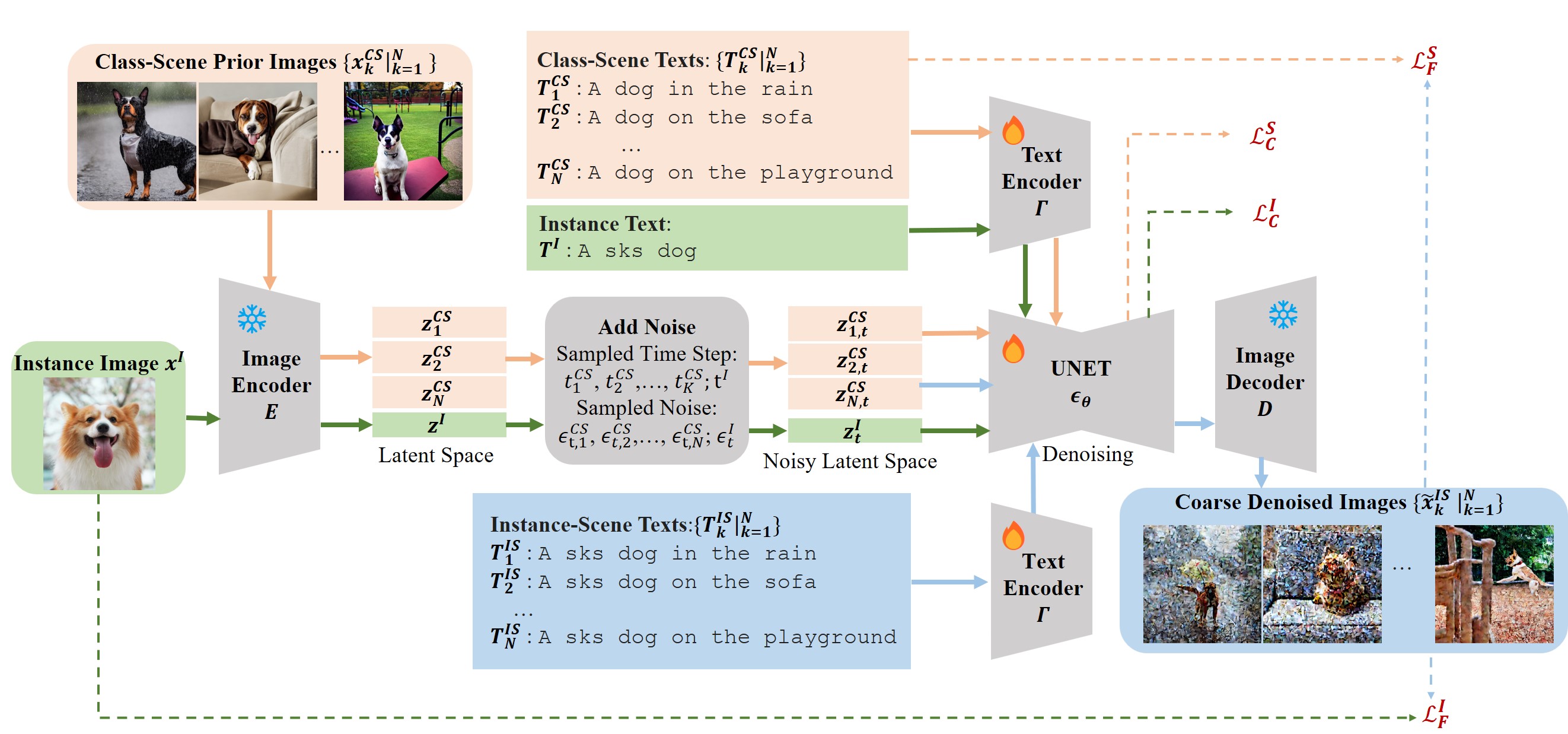}
\end{center}
\vspace{-16pt}
\caption{
The illustration of ComFusion finetuning framework. We show an example of one-shot personalized generation setting and please note ComFusion can be applied to few-shot settings. 
ComFusion consists of a \textit{composite stream} (highlighted with green and orange arrows, details in  \cref{sec:finetune} and  \cref{sec:composite}) and a \textit{fusion stream} (highlighted with blue arrows, details in  \cref{sec:fuse}).
}
\vspace{-8pt}
\label{fig:framework} 
\end{figure*}


\section{ComFusion} \label{sec:method}
In this section, we introduce ComFusion, our novel method designed to facilitate personalized subject generation across various specific scenes. Formally, with a limited number of subject instance images, typically no more than five, our objective is to synthesize new images of the subject with high detail fidelity, guided by text prompts. These prompts drive variations such as changes in the subject's location, background, pose, viewpoint, and other contextually relevant transformations. It's important to note that there are no constraints on the capture settings of the subject instance images; they can represent a wide range of scenarios.
IIn the main body of our paper, we focus on the particularly challenging \textit{one-shot} setting, where only \textit{a single instance image} is used. The generated images aim to be faithful (\textit{i.e.}, accurately reflect the content) both the subject instance and the text prompts, which manifests in two key aspects - \textit{instance fidelity}: ensuring visual congruence with the instance image and \textit{scene fidelity}: aligning the scenes in the newly created images with the provided prompts.
As shown in ~\cref{fig:framework}, 
we design a two-stream training strategy for ComFusion, consisting of a \textit{composite stream} supervised by \textit{instance finetune loss} and \textit{class-scene prior loss} (denoted as $\{\mathcal{L}_{C}^I, \mathcal{L}_{C}^S\}$ in green and orange stream of ~\cref{fig:framework}, also demonstrated in  \cref{sec:composite}) and a \textit{fusion stream} supervised by \textit{visual-textual matching loss} (denoted as $\{\mathcal{L}_{F}^I, \mathcal{L}_{F}^S\}$ in blue stream of ~\cref{fig:framework}, also demonstrated in  \cref{sec:fuse}). 

\subsection{Finetuning Text-to-Image Diffusion Models} \label{sec:finetune}
ComFusion will finetune specific pretrained diffusion models, \textit{e.g.}, Stable Diffusion~\cite{rombach2022high}, which consists of an auto-encoder (encoder $E$ and a decoder $D$), text-encoder $\Gamma$, and a denoising model $\epsilon_{\theta}$ architectured with $UNET$~\cite{ronneberger2015u}. 
The auto-encoder subjects at map an image $\bm{x}$ into low-dimensional latent $\bm{z}$ with encoder $E$ and recover the original image $\tilde{x}$ with decoder $D$ after the denoising process. 
The denoising model $\epsilon_{\theta}$ is trained on the latent to produce subject latent based on the textual condition source from $\Gamma(\bm{T})$, where $\bm{T}$ is the user-provided prompt providing the information (\textit{e.g.}, subject classes, instance attributes, scenes) of the generated images and $\Gamma$ denotes the pretrained CLIP text encoder \cite{radford2021learning}. 
Given a single subject \textit{instance image} $\bm{x}^I$ from a \textit{subject class}, the instance image is captioned with \textit{instance text} $\bm{T}^I$ = ``\texttt{a [identifier] [class noun]}'' (\emph{e.g.,} ``\texttt{a sks dog}''), where ``\texttt{[class noun]}'' is a coarse class (\emph{e.g.,} ``\texttt{dog}'') provided by user and ``\texttt{[identifier]}'' is an unique identifier for subject concept (\emph{e.g.,} ``\texttt{sks}'').  
Given a single instance image $\bm{x}^I$, the pretrained models will be finetuned with \textit{instance finetune loss}: 
\begin{equation}\label{eqn:fintune}
\begin{aligned}
&\mathcal{L}_{C}^I  = \mathbb{E}_{\bm{z} \sim \{\bm{z}^I\}, \bm{\epsilon}, t} \left[ \| \bm{\epsilon} -\epsilon_{\theta} (\bm{z}_t, t, \Gamma(\bm{T}^I)) \|_{2}^2 \right],
\end{aligned}
\end{equation}
where $t\sim\mathcal{N}(0,1)$ is the time step, $\bm{\epsilon}\sim\mathcal{N}(0,I)$ denotes the unscaled noise sampled from Gaussian distribution,  and $\bm{z}_t$ are the noisy latent at time $t$, $\{\bm{z}^I\}$  represents latent of instance images $\{\bm{x}^I\}$ processed by encoder $E$. 
This finetuning process enables the ``implantation'' of a new (unique identifier, subject) pair into the diffusion model’s ``dictionary''. The fundamental aim is to leverage the model’s existing class knowledge and intertwine it with the embedding of the subject’s unique identifier. This approach enables the effective utilization of the model’s visual priors for generating novel variations of the subject in diverse contexts.

The finetuning of diffusion models brings about the challenge of \textit{language drift} \cite{lu2020countering,lee2019countering}, a phenomenon where models gradually stray from the syntactic and semantic complexities of language, focusing too intently on task-specific details.
In our research, this issue emerges prominently: the finetuned model tends to lose the knowledge it acquired before finetuning, such as comprehending various classes and scenes that are inherent to pretrained models. This often leads to a significant reduction in \textit{scene fidelity}.
To tackle this, existing methods employ a specific \textit{prior loss} to regularize the model. 
Typically, this loss function involves using designated \textit{prior texts} $\{\bm{T}^P_i|_{i=1}^{N}\}$, input into the pretrained model to generate \textit{prior images} $\{\bm{x}^P_i|_{i=1}^{N}\}$ based on prior texts, where $N$ is the number of prior text-image pairs. 
Such a loss function ensures the model’s adherence to its pre-trained knowledge base, thus preserving essential foundational knowledge before embarking on few-shot finetuning.
An example or prior loss is class-specific prior loss proposed by DreamBooth \cite{ruiz2023dreambooth}. 
It leverages coarse class to form \textit{class text} $\bm{T}^C = $ ``\texttt{a [class noun]}'' (\emph{e.g.,} ``\texttt{a dog}''), which is fed into the pretrained model to produce \textit{class prior images} $\{\bm{x}^C_i|_{i=1}^{N}\}$ of the coarse class.
Those class prior images resemble the subject class images but has no requirement of subject instance preservation. 
Given class prior images $\{\bm{x}^C_i|_{i=1}^{N}\}$, the pretrained models will be finetuned with: 
\begin{equation}\label{eqn:DreamBooth}
\begin{aligned}
&\mathcal{L}_{dream}  = \mathbb{E}_{ \bm{z} \sim \{ \bm{z}^C\},\bm{\epsilon}, t} \left[ \| \bm{\epsilon}- \epsilon_{\theta} (\bm{z}_t, t, \Gamma(\bm{T}^C)) \|_{2}^2 \right] ,
\end{aligned}
\end{equation}
where $\{\bm{z}^C\}$ represents latent of class prior images $\{\bm{x}^C\}$ and the other terms are defined similar to \cref{eqn:fintune}. 
The class-specific prior-preservation loss supervises the model with reconstruction of class prior images, and will be trained together with \cref{eqn:fintune} with a weight hyperparameter. 
This loss function is specifically tailored to preserve the unique attributes of instance images related to their class. It utilizes semantic priors associated with the class, integrated into the model’s structure, to facilitate the generation of varied instances within the subject's class. 
Intuitively, models trained with $\mathcal{L}_{C}^I$ and $\mathcal{L}_{dream}$, leveraging the knowledge from large-scale T2I pretrained models capable of generating images with any scenes, should ideally preserve both instance and scene fidelity.
However, this approach primarily addresses language drift related to the subject class but may overlook drifts in text prompts describing the scenes of the generated images, leading to \textit{catastrophic neglecting} \cite{chefer2023attend}.
In large-scale pretrained T2I models like Stable Diffusion \cite{rombach2022high}, models trained on a vast array of image-text pairs demonstrate proficiency in generating novel images based on combinations of random texts.
Nonetheless, the neglecting phenomenon remains an issue in certain scenarios, where some prompts or subjects are not adequately generated or integrated by these large-scale models \cite{chefer2023attend}. 
In contrast, few-shot learning paradigms, relying on a limited set of image-text pairs, often yield less optimal responses to complex subject instances and scene texts than their large-scale counterparts, potentially exacerbating the neglecting issue.
This can lead to low diversity in outputs and loss to scene fidelity or adequately respond to instance images for personalized generation, as illustrated in ~\cref{fig:motivation}.

\subsection{Composite: Class-Scene Prior Loss} \label{sec:composite}
Given our objective to \textit{composite} new subject instance images within various specific scenes, we updated the prior-preservation loss in \cref{eqn:DreamBooth} to \textit{class-scene prior loss}. This update is specifically designed to maintain the pretrained model’s knowledge of both class and scene, thereby significantly enhancing \textit{scene fidelity}. 
By integrating class-scene prior loss with instance finetune loss, ComFusion effectively preserves and leverages the extensive understanding of class and scene inherent to the pretrained model.
To elaborate, we initially generate a descriptive \textit{class-scene text set} $\{\bm{T}^{CS}\}$.
This set \textit{composites} the subject class information ``\texttt{[class noun]}'' (\emph{e.g.,} ``\texttt{dog}'') and scene information ``\texttt{[scene]}'' (\emph{e.g.,} ``\texttt{in the rain}''), resulting in class-scene texts ``\texttt{a [class noun] [scene]}'' (\emph{e.g.,} ``\texttt{a dog in the rain}'').
The \textit{class-scene prior images} $\{(\bm{x}^{CS}\}$ is generated by pretrained model with the corresponding $\{\bm{T}^{CS}\}$.
Subsequently, these richly detailed class-scene image-text pairs $(\bm{x}^{CS}_k, \bm{T}^{CS}_k)$ combined with instance image-text pairs $(\bm{x}^I, \bm{T}^I)$ are fed into ComFusion to finetune the diffusion model. 
In a manner akin to \cref{eqn:DreamBooth}, the trainable parameters of ComFusion are optimized by class-scene prior loss:
\begin{equation}\label{eqn:composite_loss}
\begin{aligned}
&\mathcal{L}_{C}^S  =\mathbb{E}_{ (\bm{z}, \bm{T}) \sim \{(\bm{z}^{CS},  \bm{T}^{CS})\},\bm{\epsilon}, t} \left[ \| \bm{\epsilon} - \epsilon_{\theta} (\bm{z}, t, \Gamma(\bm{T})) \|_{2}^2 \right] ,
\end{aligned}
\end{equation}
where $\{(\bm{z}^{CS},  \bm{T}^{CS})\}$ is the set of latent-text pairs corresponding to $\{(\bm{x}^{CS}, \bm{T}^{CS})\}$.
Similar to $\mathcal{L}_{dream}$ in \cref{eqn:DreamBooth}, this class-scene prior loss will be trained together with instance finetune loss $\mathcal{L}_{C}^I$ in \cref{eqn:fintune} with a $\lambda_C$ controls for the relative weight of this term. 
$\mathcal{L}_{C}^I$ and $\mathcal{L}_{C}^S$ formulate the objective of the \textit{composite stream} in ComFusion.
Different from $\mathcal{L}_{dream}$, the class-scene prior text $\bm{T}^{CS}$ in $\mathcal{L}_{C}$ articulates a comprehensive delineation of subject class information and meticulous scene descriptors as specified by $\bm{T}^{CS}$.
This loss formulation adeptly tackles the language drift issue related to class and scene knowledge in the finetuned model. It facilitates the generation of a varied collection of images that capture the essence of both class and scene priors from the pretrained model, while integrating these elements with subject instances and their specific contexts. Such integration enables ComFusion to produce images that accurately depict subject instances within their respective scenes.
\begin{figure}[htp]
\begin{center}
\includegraphics[width=1.0\linewidth]{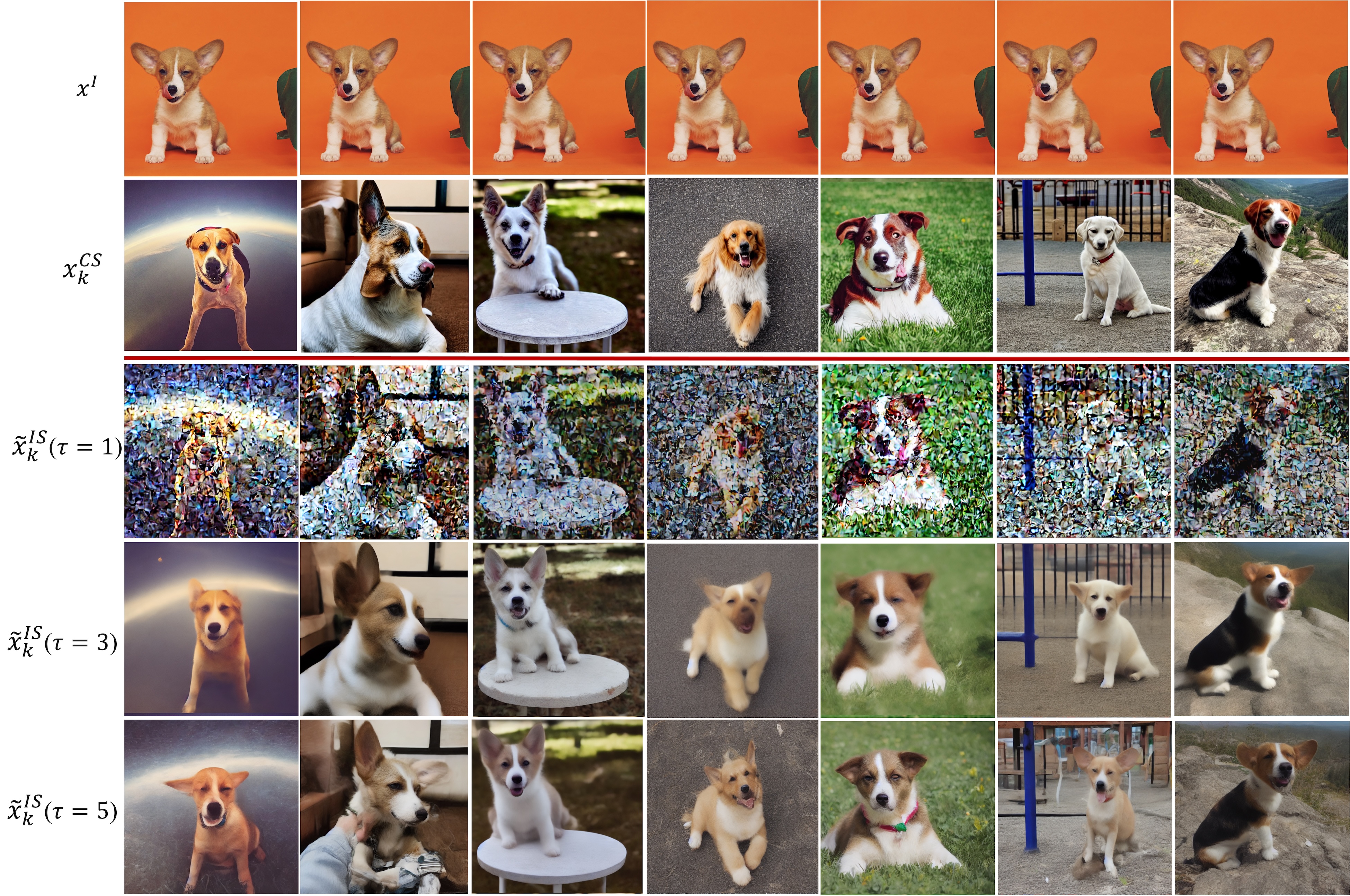}
\end{center}
\vspace{-16pt}
\caption{The coarse generated results $\tilde{\bm{x}}^{IS}_k$ by supervision of visual-textual fusion loss $\{\mathcal{L}_F^I,\mathcal{L}_F^S\}$ under denoising steps $\tau$ sampled from $\{1,3,5\}$. The instance in coarse generated images is similar to instance image $\bm{x}^I$ , while it maintains consistent with specific scene in prior images $\bm{x}^{CS}_k$.}
\vspace{-8pt}
\label{fig:process} 
\end{figure}

\subsection{Fusion: Visual-Textual Matching Loss}\label{sec:fuse}
In ComFusion, the composite stream effectively finetunes the instance image while retaining the class-scene prior knowledge embedded in the pretrained model. This approach effectively maintains both instance fidelity and scene fidelity. 
Furthermore, ComFusion employs a \textit{visual-textual matching loss} to \textit{fuse} the visual information of the subject instance with the textual context of the scene, further enhancing both instance fidelity and scene fidelity. This ensures a more coherent and accurate depiction of the subject instance and the scenes.
The key idea behind the visual-textual matching loss is to generate coarse images that encapsulate both the subject instance and scene texts. 
This loss then guarantees that the coarsely generated images effectively merge the specific visual details of the instances with the textual nuances of the scenes, aligning them with the instance image and scene texts.
Specifically, for a class-scene prior image $\bm{x}^{CS}$ annotated with detailed class-scene text $\bm{T}^{CS}$, according to the standard forward process of DDPMs~\cite{ho2020denoising}, we add a random and appropriate level of noise $\bm{\epsilon}_t$ with timestep $t$ to obtain noisy latent $\bm{z}^{CS}_t$. This is designed to infuse new information while preserving the structural features of class-scene prior images. 
We then create \textit{instance-scene text} $\bm{T}^{IS}$ by replacing the ``\texttt{[class noun]}'' with ``\texttt{[identifier] [class noun]}'' in class-scene text (\emph{e.g.,} ``\texttt{a sks dog in the rain}''). 
This modified text is processed through by the text-encoder $\Gamma$ to obtain conditional textual information, which is then used to iteratively denoise the noisy latent $\bm{z}^{CS}_t$ to denoised latent $\tilde{\bm{z}}^{IS}$. 
Adopting an accelerated generation process in DDIMs~\cite{song2020denoising}, we set $\tau$ as the number of steps required to denoise $\bm{z}^{CS}_t$ into $\tilde{\bm{z}}^{IS}$, with a function expressed as: 
\begin{equation}\label{eqn:ddim}
    \tilde{\bm{z}}^{IS} = f_{\theta}(\bm{z}^{CS}_t, \bm{T}^{IS}, \tau).
\end{equation}
Generally, $f_{\theta}$ significantly reduces the computational effort required for denoising from $t$ to $\tau$,  producing a coarse denoised latent.
\textit{The specifics of this denoising function $f_{\theta}$ in \cref{eqn:ddim} will be detailed in the supplementary materials.}
The denoised latent $\tilde{\bm{z}}^{IS}$ is then decoded using decoder $D$, resulting in a \textit{coarse denoised image} $\tilde{\bm{x}}^{IS} = D(\tilde{\bm{z}}^{IS})$.
This image, guided by the instance-scene text, fuses both the subject and the scene's features. \cref{fig:process} illustrates examples of coarse denoised images under different settings of $\tau$.
We enable the instance image $\tilde{\bm{x}}^{IS}$ to have similar textural structure (\emph{resp.,} visual appearance) of class-scene prior image $\bm{x}^{CS}$ (\emph{resp.,} instance image $\bm{x}^I$) by a pair of visual-textual fusion loss:
\begin{equation}\label{eqn:fusion_loss}
\begin{aligned}
&\mathcal{L}_{F}^I  = \mathbb{E}_{ \bm{x} \sim \{ \tilde{\bm{x}}^{IS}_k\} } \left[  -  \textbf{DINO}(\bm{x} , \bm{x}^I)  \right], \\
&\mathcal{L}_{F}^S  =\mathbb{E}_{ (\bm{x}', \bm{T}) \sim \{ (\tilde{\bm{x}}^{IS}_k,  \bm{T}^{CS}_k)\} } \left[-  \textbf{CLIP}(\bm{x}', \bm{T})  \right], 
\end{aligned}
\end{equation}
where the first (\emph{resp.,} second) term is represented by $\mathcal{L}^I_{F}$ (\emph{resp.,} $\mathcal{L}^S_{F}$) in ~\cref{fig:framework}.
DINO~\cite{caron2021emerging} is a self-supervised pretrained transformer, renowned for its proficiency in extracting visual information from images. Utilizing self-supervised learning techniques, DINO effectively discerns and encodes complex visual patterns, making it exceptionally suitable for image analysis tasks.
In contrast, CLIP~\cite{radford2021learning} is at the forefront of image-text cross-modality pretraining. It is designed to align visual information with corresponding textual data, thus enabling the model to comprehend and relate the contents of images with their relevant textual descriptions. CLIP's ability to bridge visual and textual domains renders it an invaluable asset for tasks that demand a thorough understanding of both visual and textual elements, facilitating precise and effective image-text alignments.
Hence, in the fusion stream of our approach, we employ DINO for visual matching and CLIP for textual matching, leveraging the strengths of each model to enhance the overall efficacy of our method.
$\textbf{DINO}(\tilde{\bm{x}}^{IS}_k , \bm{x}^I)$ is used to calculate the cosine similarity between DINO embedding of $\tilde{\bm{x}}^{IS}_k$ and $\bm{x}^I$ with pretrained ViT-S/16 DINO~\cite{caron2021emerging}, the larger similarity means that generated $\tilde{\bm{x}}^G_k$ is more similar to instance image $\bm{x}^I$.
$\textbf{CLIP}(\tilde{\bm{x}}^{IS}_k, \bm{T}^{CS}_k)$ aims at calculating the cosine similarity of visual embedding of generated image $\tilde{\bm{x}}^{IS}_k$ and textual embedding of class-scene prior text $\bm{T}^{CS}_k$.
To the larger similarity, the trained model tend to produce new images including specific scene information. 
By applying both visual loss $\mathcal{L}^I_{F}$ and textual loss $\mathcal{L}^S_{F}$ on the same generated image $\tilde{\bm{x}}^{IS}_k$, ComFusion mitigates the catastrophic neglecting problem and achieves a better balance between instance fidelity and scene fidelity. 
As a result, it yields a more harmonious and precise depiction that effectively captures the core characteristics of instance fidelity and scene fidelity.

\subsection{Overall Objectives and Inference Process}

ComFusion's objective function integrates the instance finetune loss in \cref{eqn:fintune} is combined with the class-scene prior loss in \cref{eqn:composite_loss} and visual-textual fusion loss in \cref{eqn:fusion_loss}:
\begin{equation}\label{eqn:total_loss}
\begin{aligned}
\mathcal{L}_{total}  =  \mathcal{L}_C^I + \lambda_{C}^{S} \mathcal{L}_C^S  + \lambda_{F}^{I} \mathcal{L}_F^I + \lambda_{F}^{S} \mathcal{L}_F^S,  
\end{aligned}
\end{equation}
where $\lambda_{C}^{S}$, $\lambda_{F}^{I}$, and $\lambda_{F}^{S}$ represent the respective weights of $\mathcal{L}^S_{C}$, $\mathcal{L}^I_{F}$, and $\mathcal{L}^S_{F}$.
$\mathcal{L}_{total}$  is employed to finetune the trainable parameters of text-encoder $\Gamma$ and UNET $\epsilon_{\theta}$ based on pretrained Stable Diffusion~\cite{rombach2022high}. 
During this process, the parameters of the auto-encoders remain fixed.
In the inference phase, ComFusion follows the standard T2I inference protocol: generating a random latent, followed by denoising this latent using the prompt ``\texttt{a [identifier] [class noun] [scene]}'' with the UNet. Finally, the denoised latent is decoded to produce new images.

\section{Experiments}

\subsection{Experimental Settings and Details}
\noindent\textbf{Implementation Details.} 
All methods were applied using a pre-trained Stable Diffusion (SD) checkpoint 1.5~\cite{rombach2022high}. 
We trained ComFusion and DreamBooth for 1200 steps, using a batch size of 1 and learning rate $1 \times 10^{-5}$. The number of prior images $N$ is set as $200$ for fair comparison. 
During training, the hyper parameters $\lambda_C^S$ (\emph{resp.,} $\lambda_F^S$, $\lambda_F^I$,$\tau$) is set as $1$ (\emph{resp.,} $0.01$, $0.01$,$3$). All experiments are conducted with $1$ $A100$ GPU.
Detailed implementation information for all baselines is provided in Supplementary.

\noindent\textbf{Datasets.} To evaluate the effectiveness of our proposed methods among different datasets, we use a combined dataset of the TI~\cite{gal2022image} dataset of 5 concepts, and the dataset from DreamBooth~\cite{ruiz2023dreambooth} with 20 concepts. For both datasets, each concept selects one original image as instance image. We perform experiments on $25$ subject datasets spanning a variety of categories and varying training samples. We evaluate all the methods with $15$ distinct scenes. Also, we use $\bm{T}^{CS} = $ ``\texttt{a [class noun] Scene}'' with the same scene prompts to sample prior images with $15$ scenes for ComFusion. Detailed information about the subject datasets and scene prompts is available in the supplementary materials. Experiments involving more than one instance image, other specific scenes, and scenarios without specific scenes for all methods are also documented in Supplementary.

\noindent\textbf{Baselines.}  We compare our ComFusion described in Section~\cref{sec:method} with DreamBooth~\cite{ruiz2023dreambooth}, Textual-Inversion(TI)~\cite{gal2022image}, Custom-Diffusion (CD)~\cite{kumari2023multi}, Extended Textual-Inversion (XTI)~\cite{voynov2023p+}, ELITE~\cite{wei2023elite}, and Break-A-Scene~\cite{avrahami2023break}. Details of these baseline methods are reported in Supplementary.

\noindent\textbf{Evaluation Metrics.} 
Following DreamBooth~\cite{ruiz2023dreambooth}, for each method, we generated $10$ images for each of $25$ instances and each of $15$ scenes, totaling $3750$ images for evaluation of robustness and generalization abilities of each method.
Following DreamBooth~\cite{ruiz2023dreambooth} and CD~\cite{kumari2023multi},  we evaluate those methods on two dimensions including instance fidelity and scene fidelity.
CLIP-I~\cite{radford2021learning} and DINO score~\cite{caron2021emerging} were used to evaluate instance fidelity by measuring the similarity between generated images and instance images, and the alignment between textual scene with generated images are measured by CLIP-T~\cite{radford2021learning}.
Detailed descriptions of these measurement metrics are provided in Supplementary.

\begin{table}[!htp]
  \centering
  \caption{Quantitative metric comparison of instance fidelity (DINO, CLIP-I) and scene fidelity (CLIP-T).} 
  \vspace{-8pt}
   \resizebox{0.9\columnwidth}{!} {
  \begin{tabular}{l|ccc}
\toprule 
Methods&  DINO ($\uparrow$)& CLIP-I ($\uparrow$)&  CLIP-T ($\uparrow$)\\
\midrule 
Real Images&\textit{0.795} & \textit{0.859}&N/A \\
\midrule
DreamBooth~\cite{ruiz2023dreambooth}&0.619&0.752& 0.229 \\
TI~\cite{gal2022image}&0.465&0.634& 0.185 \\
CD~\cite{kumari2023multi}&0.615&0.724 & 0.205\\
XTI~\cite{voynov2023p+}&0.435&0.601& 0.198 \\
ELITE~\cite{wei2023elite} &0.405&0.615&0.249 \\
Break-A-Scene~\cite{avrahami2023break}&0.632&0.771 & 0.294 \\
\midrule
Ours& \bf{0.658} & \bf{0.814} & \bf{0.321} \\
\bottomrule
\end{tabular}
}

  \label{tab:metric_score}
\end{table}

\begin{table}[!htp]
  \centering
  \caption{Instance fidelity and scene fidelity user preference.} 
  \vspace{-8pt}
  \resizebox{0.9\columnwidth}{!} {
  \begin{tabular}{l|cc}
\toprule 
Methods&  Instance fidelity ($\uparrow$) &  Scene fidelity ($\uparrow$) \\
\midrule 
DreamBooth~\cite{ruiz2023dreambooth}& 3.1\% & 3.8\% \\
TI~\cite{gal2022image}&0.3\% & 0.0\%\\
CD~\cite{kumari2023multi}&6.2\% & 1.0\%\\
XTI~\cite{voynov2023p+}& 0.3\% & 1.8\%\\
ELITE~\cite{wei2023elite} & 0.0\% & 11.1\%  \\
Break-A-Scene~\cite{avrahami2023break}&34.5\% & 20.2\% \\
\midrule
Ours &55.6\%& 62.1\%\\
\bottomrule
\end{tabular}
}

  \label{tab:user_study_score}
\end{table}
\begin{figure*}[htp]
\begin{center}
\includegraphics[width=1.0\linewidth]{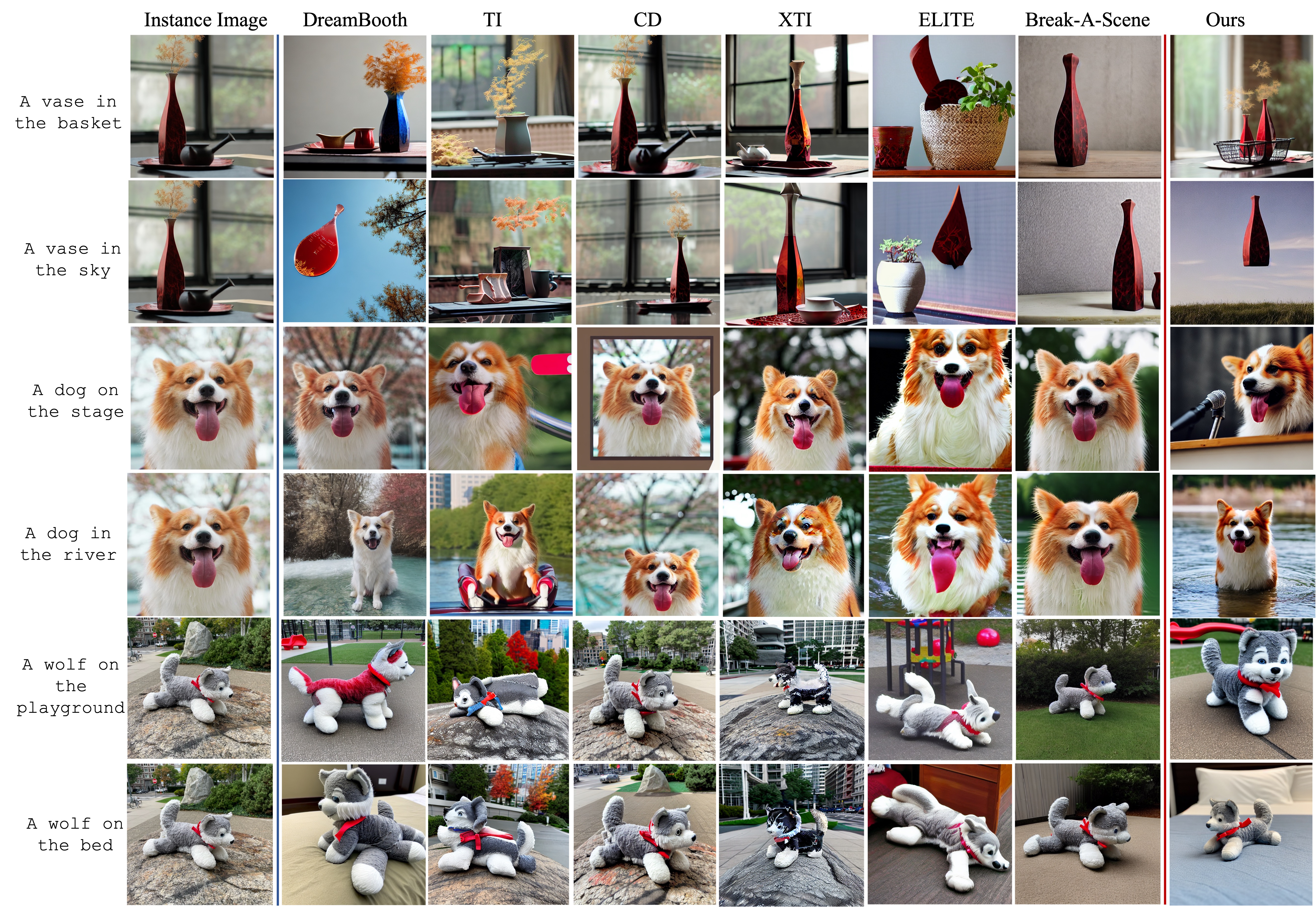}
\end{center}
\vspace{-16pt}
\caption{Images generated by DreamBooth~\cite{ruiz2023dreambooth}, TI~\cite{gal2022image},CD~\cite{kumari2023multi},XTI~\cite{voynov2023p+},ELITE~\cite{wei2023elite}, Break-A-Scene~\cite{avrahami2023break}, and our proposed ComFusion in multiple specific scenes from a single instance image.}
\vspace{-8pt}
\label{fig:comparison} 
\end{figure*}


\subsection{Comparisons with Personalized Generation Baselines}
We perform quantitative and qualitative evaluation on the instance fidelity and scene fidelity of generated images. Instance fidelity assesses how well the generated image maintains the identity of the original instance image, while scene fidelity evaluates the semantic similarity between the generated image and the input textual scene.

\subsubsection{Quantitative Evaluations}

\noindent\textbf{Quality Assessments of Generated Images.}
We perform the quantitative evaluation on the instance fidelity using DINO score and CLIP-I score, and scene fidelity with CLIP-T score. 
In ~\cref{tab:metric_score}, ``Real Images'' represents measure the similarity between given single image and remaining real images belonging to the same subject as given image, providing upper bound of fidelity of subject. Comparisons in ~\cref{tab:metric_score} indicate that our ComFusion achieves the highest scores for DINO, CLIP-I, and CLIP-T, indicating that it can generate high-fidelity images with higher instance fidelity and scene fidelity than baseline methods. 

\noindent\textbf{Human Perceptual Study.}
Further, following DreamBooth~\cite{ruiz2023dreambooth}, we conduct a user study to evaluate the instance fidelity and scene fidelity of generated images. In detail, based on generated $3750$ images per method including $6$ baseline methods and our method, we present results generated from different methods in random order and we ask 12 users to choose.  
(1) Instance fidelity: determining which result better preserves the identity of the instance image, and (2) Scene fidelity: evaluating which result achieves better alignment between the given textual scene and the generated image. 
We collect 90,000 votes from 12 users ($12 \times 3750 \times 2$) for instance fidelity and scene fidelity, and show the percentage of votes for each method in ~\cref{tab:user_study_score}. 
The comparison results demonstrate that the generated results obtained by our method are preferred more often than those of other methods.

\subsubsection{Qualitative Evaluations}
To evaluate the superiority of our ComFusion in balancing the accuracy of subjects and the consistency of multiple specific scenes, we visualize comparison results in ~\cref{fig:comparison}. We can see that images generated by TI~\cite{gal2022image}, CD~\cite{kumari2023multi}, and XTI~\cite{voynov2023p+} are similar to input instance image in terms of structure, those methods fail to make response to the specific scene in given testing prompts. ELITE~\cite{wei2023elite} may generate distorted images in unexpected scenes. Images generated by Break-A-Scene~\cite{avrahami2023break} maintain instance fidelity while may fail to composite subject instance in specific scenes. 
In contrast, our ComFusion can generate images of higher instance fidelity and scene fidelity. This is attributed to the class-scene prior loss can introduce specific scene information during the process of learning subject instance and visual-textual matching loss can enhance the fusion between visual instance image and textual scene context.  

\subsection{Ablation Studies}

\noindent\textbf{Effect of Class-Scene Prior Loss}
Compared with prior preservation loss (\cref{eqn:DreamBooth}) proposed in DreamBooth~\cite{ruiz2023dreambooth}, our class-scene prior loss $\mathcal{L}_C^S$ (\cref{eqn:composite_loss}) utilizes detailed texts for prior images to incorporate multiple specific scenes. During training, this loss explicitly enforces the model to retain prior scene knowledge while incorporating new information from instance images within these scenes.  
From the visual comparison between ComFusion(``w/o visual-textual loss $\{\mathcal{L}_F^I, \mathcal{L}_F^S\}$'') and DreamBooth~\cite{ruiz2023dreambooth} in ~\cref{fig:motivation}, ~\cref{fig:comparison}, and quantitative results in ~\cref{tab:ablation}, we can see that class-scene prior loss $\mathcal{L}_C^S$ significantly improves the CLIP-T score while achieves comparable CLIP-I(\emph{resp.,} DINO) score, which indicates that it can effectively improve scene fidelity without undermining the instance fidelity.  
 
\begin{table}[!htp]
  \centering
  \caption{Ablation studies of each loss and alternative designs. Time represents the totoal training time on 1 A100 GPU.} 
  \vspace{-8pt}
  \resizebox{0.9\columnwidth}{!} {
  \begin{tabular}{l|ccc|c}
\toprule 
Methods&  DINO ($\uparrow$)& CLIP-I ($\uparrow$)&  CLIP-T ($\uparrow$) & Time(/s)\\
\midrule 
Real Images&\textit{0.795} & \textit{0.859}&N/A & N/A\\
\midrule
DreamBooth &0.619&0.752& 0.229 & 491.8 \\
\hline
Ours  (w/o $\{\mathcal{L}_F^I, \mathcal{L}_F^S\}$)&0.627&0.771&0.301& 597.0\\
Ours (w/o $\mathcal{L}_F^I$) &0.586&0.697&0.342& 597.3\\
Ours  (w/o $\mathcal{L}_F^S$) &0.716&0.828&0.189 & 597.1\\
\hline
Ours ($\tau=1$) &0.641&0.806&0.334& 537.9 \\
Ours ($\tau=5$) &0.698&0.825&0.309& 623.1 \\
\midrule
Ours ($\tau=3$)& 0.658 & 0.814 & 0.321 & 597.7 \\
\bottomrule
\end{tabular}
}

  \label{tab:ablation}
\end{table}

\begin{figure}[htp]
\begin{center}
\includegraphics[width=1.0\linewidth]{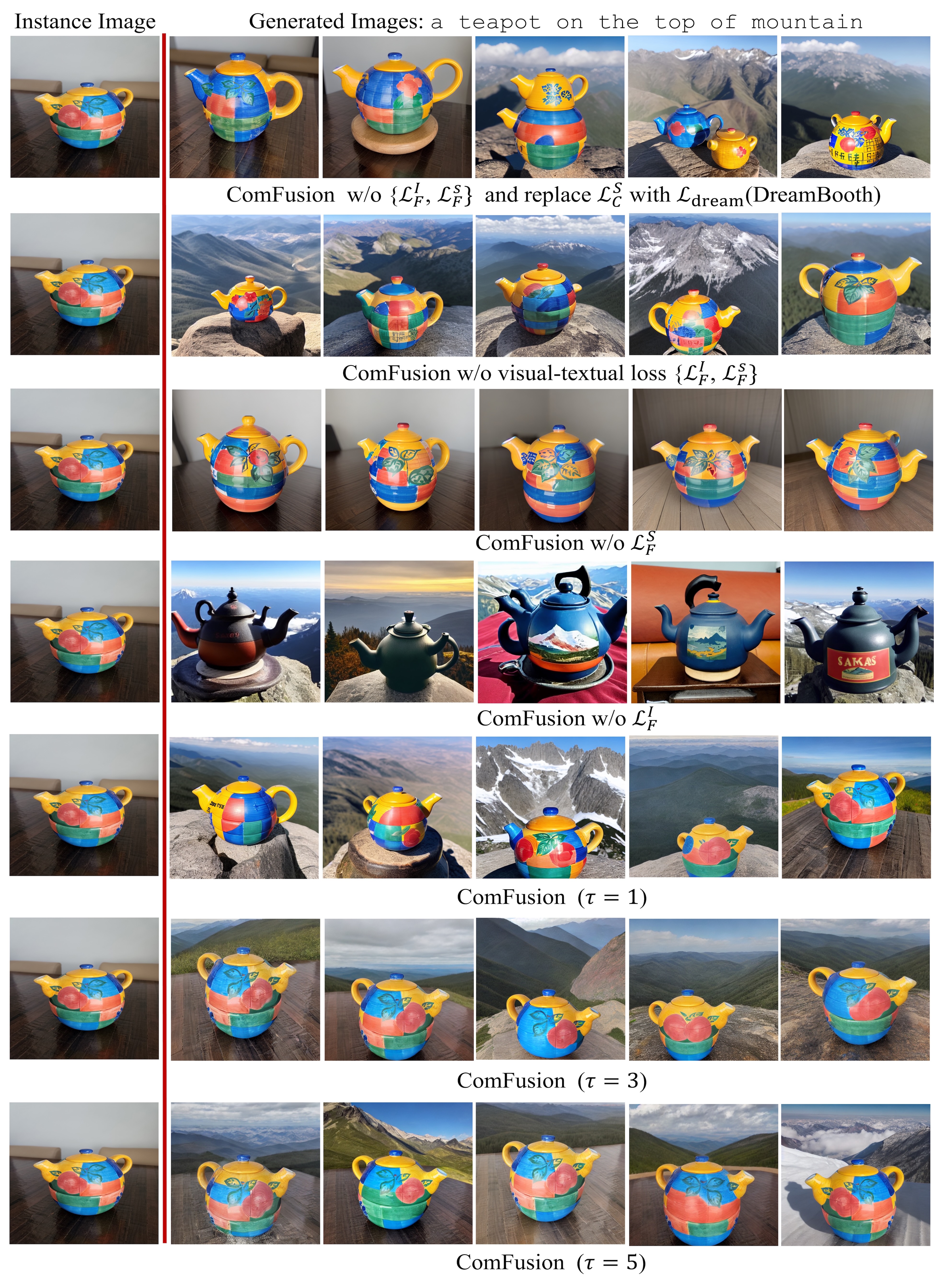}
\end{center}
\vspace{-16pt}
\caption{Visual ablative results of ComFusion.}
\vspace{-8pt}
\label{fig:ablation} 
\end{figure}

\noindent\textbf{Effect of Visual-Textual Matching Loss}
We further conduct ablation to evaluate the effect of the proposed visual-textual loss $\{\mathcal{L}_F^I, \mathcal{L}_F^S\}$ in \cref{eqn:fusion_loss}. To evaluate the effect of each item in visual-textual loss, we alternatively remove $\{\mathcal{L}_F^I, \mathcal{L}_F^S\}$ (\emph{resp.,}$\mathcal{L}_F^I$, $\mathcal{L}_F^S$) from total loss function in \cref{eqn:total_loss}, and report visual results in ~\cref{fig:ablation} and quantitative results in ~\cref{tab:ablation}. The comparison results indicate that $\{\mathcal{L}_F^I, \mathcal{L}_F^S\}$ are well-design to balance the instance fidelity and scene fidelity, removing them degrades the instance fidelity in generated images in 2nd row in ~\cref{fig:ablation}. To further study effect of each item in visual-textual loss, removing $\mathcal{L}_F^S$ leads to degrading the score of DINO and CLIP-I reflecting by poor instance fidelity in 4th row in ~\cref{fig:ablation}, while lower scene fidelity in 3rd row in  ~\cref{fig:ablation} is caused by removing $\mathcal{L}_F^C$. 

\noindent\textbf{Effect of Denoising Timesteps $\tau$}
To assess the impact of timesteps $\tau$ in the fusion stream, we experimented with varying $\tau$ values from $\{1,3,5\}$ to train ComFusion. Our observations indicate that a larger $\tau$ value tends to better preserve instance fidelity but at the expense of reduced scene fidelity. We selected $\tau=3$ as the default setting for ComFusion, considering time cost and a balance between instance fidelity and scene fidelity




\section{Conclusions}
In this paper, we present ComFusion, a novel approach designed to facilitate personalized subject generation within multiple specific scenes from a single image.
ComFusion introduces a class-scene prior loss to composite knowledge of subject class and specific scenes from pretrained models. 
Moreover, a visual-textual matching loss to further improve the fusion of visual object feature and textual scene feature.
Extensive quantitative and qualitative experiments demonstrate the effectiveness of ComFusion.

{\small
\bibliographystyle{ieeenat_fullname}
\bibliography{egbib}
}


\end{document}




\appendix
\section*{\Large Appendix}

In this supplementary material, we provide additional material to complement our main submission. Key sections are outlined as follows: 
In \cref{sec:f_theta}, we elucidate  the background and detailed methodology of the coarse denoising function.
In \cref{sec:implementary_detail}, we provide the implementation details of the baseline methods.  
Evaluation metrics of CLIP-I, CLIP-T, and DINO used in our study are represented in \cref{sec:evaluation_metric_details}. 
In \ref{sec:dataset_introduction}, we present the specific scenes and visualize the single instance image from the training dataset. 
In \ref{sec:unseen_scene}, we evaluate  the performance of the proposed ComFusion and baseline methods in scenarios involving unseen scenes, testing the generalizability of ComFusion.
In \cref{sec:multiple_instance}, the performance of ComFusion, when trained with multiple instance images, against the DreamBooth baseline method, demonstrating its effectiveness in varied training contexts.  
In \cref{sec:visualization_supple}, we visualize additional generated images by both baseline methods and ComFusion, offering more examples of our model's capabilities. 
Finally, in \cref{sec:limitation}, discuss some failure cases in complex scenes, highlighting the current limitations and potential areas for personalized subject generation.


\section{The Coarse Denoising Function} \label{sec:f_theta}
In this section, we elucidate the background and detailed methodology of the coarse denoising function outlined in Eq. (4) from Sec. 3.3. Within the fusion stream of ComFusion, a visual-textual matching loss is employed to integrate the visual information of the subject instance with the textual context of the scene. To achieve this, we generate a coarse denoised image $\tilde{\bm{x}}^{IS}$, under the guidance of the instance-scene text $\bm{T}^{IS}$ thereby fusing the characteristics of both the subject and the scene.
The coarse denoised image $\tilde{\bm{x}}^{IS} = D(\tilde{\bm{z}}^{IS})$ is derived from the denoised latent  $\tilde{\bm{z}}^{IS}$, which in turn is computed by the denoising function $f_{\theta}(\bm{z}^{CS}_t, \bm{T}^{IS}, \tau)$ as defined in Eq. (4) of the main text. Here, $\bm{z}^{CS}_t$ represents the noisy latent at timestep $t$ from the class-scene prior image ${\bm{x}}^{CS}$ , and $\tau$ is a hyperparameter that efficiently reduces the computational demands of the denoising process.

Diffusion models \cite{sohl2015deep,ho2020denoising} have the capability to generate realistic images from a normal distribution by reversing a gradual noising process. The forward process, denoted as $q(\cdot)$, constitutes a Markov chain that incrementally transforms data from $\bm{x}_0 \sim q(\bm{x})$ to a Gaussian distribution. 
A single step in the forward process is defined as:
\begin{equation}
    q(\bm{x}_t | \bm{x}_{t-1}) = \mathcal{N}(\bm{x}_t; \sqrt{1 - \beta_{t}} \bm{x}_{t-1}, \beta_{t} I),
\end{equation} 
where $\beta_t$ represents a predefined variance schedule over $T$ steps.
The forward process allows for the sampling of $\bm{x}_t$ at any given timestamp $t$ in a closed form:
\begin{equation}
\bm{x}_t = \sqrt{\alpha_t} \bm{x}_0 + \sqrt{1 - \alpha_t} \epsilon,
\end{equation}
where
\begin{equation}
\alpha_t = \prod_{s=1}^{t} (1 - \beta_s), \quad \epsilon \sim \mathcal{N}(0, I).
\end{equation}
The reverse process in diffusion models is defined as:
\begin{equation}
    p_{\theta}(\bm{x}_{t-1} | \bm{x}_t) = \mathcal{N}(\bm{x}_{t-1}; \mu_{\theta}(\bm{x}_t, t), \sigma_{\theta}(\bm{x}_t, t)I).
\end{equation} 
This process can be parameterized using deep neural networks. Denoising Diffusion Probabilistic Models (DDPMs) \cite{ho2020denoising} have demonstrated that utilizing a noise approximation model  $\epsilon_{\theta}(\bm{x}_t, t)$ is more effective than using $\mu_{\theta}(\bm{x}_t, t)$ for procedurally transforming the prior noise into data.
As a result, the sampling in diffusion models is performed according to the following equation:
\begin{equation}
\bm{x}_{t-1} = \frac{1}{\sqrt{1 - \beta_t}} \left( \bm{x}_t - \frac{\beta_t}{\sqrt{1 - \alpha_t}} \epsilon_{\theta}(\bm{x}_t, t) \right) + \sigma_t \bm{\epsilon}.
\end{equation}

Recently, Latent Diffusion Models (LDM) \cite{rombach2022high} have been developed to reduce computational cost by operating the diffusion model within a latent space. LDM utilizes a pretrained encoder $E$ to embed an image into latent space, and a pretrained decoder $D$ for image reconstruction. In LDM, the diffusion process is defined using $\bm{z}$ ($\bm{z} = E(\bm{x})$) instead of $\bm{x}$ itself.
LDM adopts the Denoising Diffusion Implicit Models (DDIM) \cite{song2020denoising} sampling process, which is based on an Euler discretization of a neural Ordinary Differential Equation (ODE) \cite{chen2018neural}. This approach enables fast and deterministic sampling. Intuitively, the DDIM sampler directly predicts $\bm{z}_0$ directly from $\bm{z}_t$, then generates $\bm{z}_{t-1}$ through a reverse conditional distribution. 
Specifically, integrating the textual condition $\bm{T}$ and the text encoder $\Gamma(\cdot)$, $h_{\theta}(\bm{z}_t, t, \bm{T})$ is the predicted $\bm{z}_0$ given $\bm{z}_t$ and $t$:
\begin{equation} \label{eqn:predict_h}
h_{\theta}(\bm{z}_t, t, \bm{T}) = \frac{\bm{z}_t - \sqrt{1 - \alpha_t} \epsilon_{\theta}(\bm{z}_t, t, \Gamma( \bm{T}))}{\sqrt{\alpha_t}}.
\end{equation}
The deterministic sampling process in LDM using DDIM can be outlined as follows:
\begin{equation}
\bm{z}_{t-1} = \sqrt{\alpha_{t-1}} h_{\theta}(\bm{z}_t, t, \bm{T}) + \sqrt{1 - \alpha_{t-1}} \epsilon_{\theta}(\bm{z}_t, t, \Gamma( \bm{T})).
\end{equation}
Once the diffusion process is complete, the image is reconstructed by the decoder $D$, such that $\tilde{\bm{x}} = D(\bm{z})$.

The coarse denoising function $f_{\theta}(\cdot)$ is formulated based on $h_{\theta}(\cdot)$ in \cref{eqn:predict_h}. 
This function is specifically designed to generate coarse denoised images through a $\tau$ steps iteration process.
The application of the coarse denoising function is defined as follows:
\begin{equation}
f_{\theta}(\bm{z}_t, \bm{T}, \tau) = 
\left\{\begin{array}{lr}
h_{\theta}(\bm{z}_t, t, \bm{T}) & (\tau=1) \\
f_{\theta}\left(
\sqrt{\alpha_{r(\tau, t)}} h_{\theta}\left(\bm{z}_t, t, \bm{T}\right) + \sqrt{1 - \alpha_{r(\tau, t)}} \epsilon_{\theta}\left(\bm{z}_t, t, \Gamma( \bm{T})\right),  \bm{T}, \tau-1 \right) 
& (\text { o.w. })
\end{array}\right .
\end{equation}
where $r(\tau, t) = \lceil\frac{v\times t}{\tau}\rceil$.
The function $f_{\theta}(\cdot)$ is intended for recursive application, executed $\tau$ times.
Each iteration of $f_{\theta}(\cdot)$ progressively reduces the noise in $\bm{z}_t$, finally resulting in a coarse denoised latent.

In the implementation of our coarse denoising process, we strategically sample the timestep $t$ from $[\lceil0.2T\rceil, \lceil0.8T\rceil]$. This specific range is chosen to ensure that the coarse denoised image $\tilde{\bm{x}}^{IS}$ effectively fuses information from both the subject instance and the scene text.
Please note $\tilde{\bm{x}}^{IS} = D(f_{\theta}(\bm{z}^{CS}_t, \bm{T}^{IS}, \tau))$.
If $t$ is too close to $1$, the influence of the instance-scene text $\bm{T}^{IS}$ becomes limited, resulting in $\tilde{\bm{x}}^{IS}$ lacking sufficient visual cues of the subject instance.
On the other hand, if $t$ approaches $T$, the effect of the class-scene latent $\bm{z}^{CS}$ diminishes due to excessive noise, causing a loss of scene information in $\tilde{\bm{x}}^{IS}$.
To mitigate these issues and achieve a balanced fusion of instance and scene features, we opt for sampling $t$ within the middle range of $[\lceil0.2T\rceil, \lceil0.8T\rceil]$.

\begin{figure*}[htp]
\begin{center}
\includegraphics[width=1.0\linewidth]{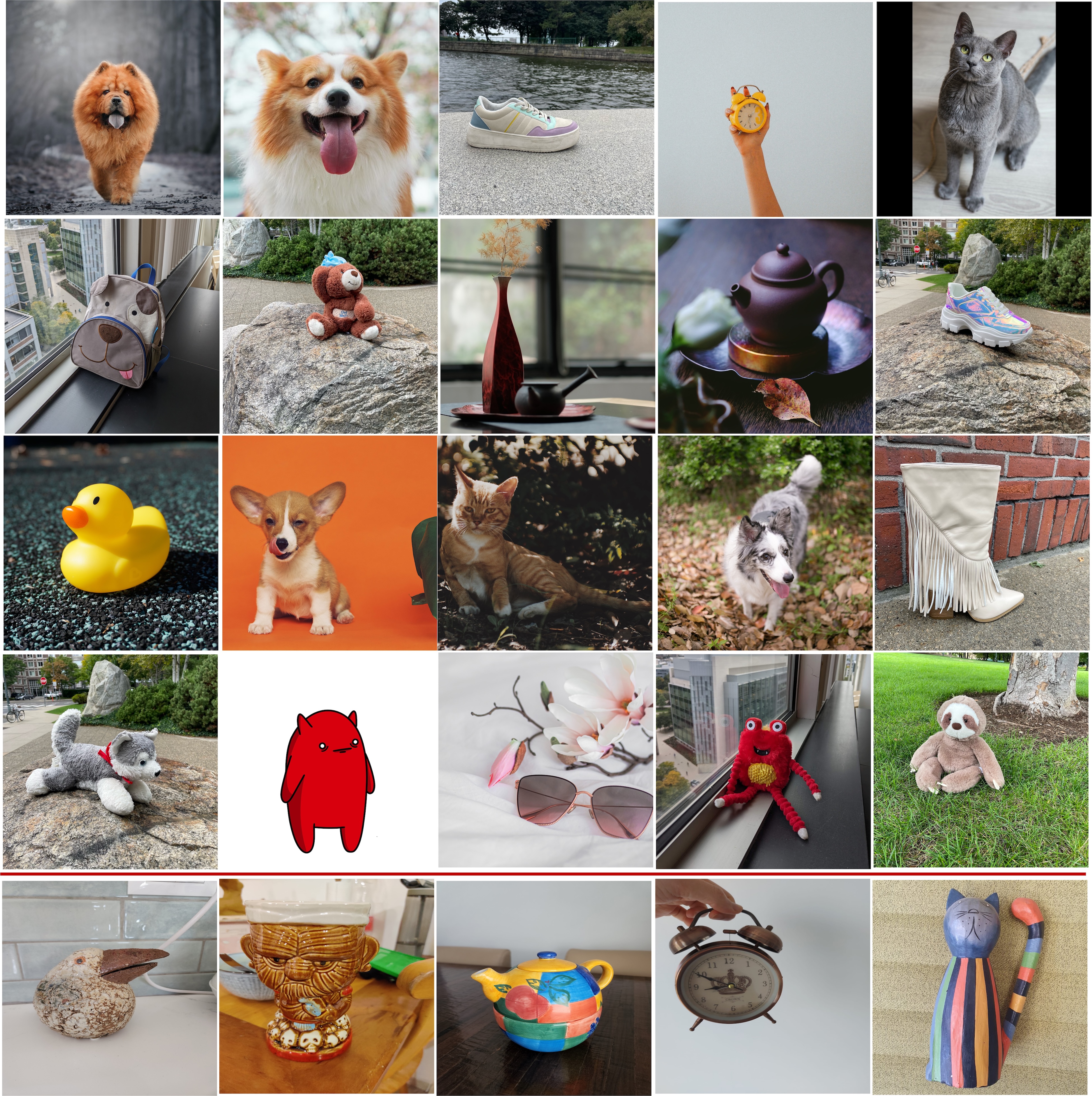}
\end{center}
\vspace{-16pt}
\caption{25 concept image from DreamBooth~\cite{ruiz2023dreambooth} and TI~\cite{gal2022image} datasets. The images in last row from TI~\cite{gal2022image} dataset and others from DreamBooth~\cite{ruiz2023dreambooth} dataset.}
\label{fig:datasets} 
\end{figure*}

\section{Implementation Details} \label{sec:implementary_detail} 

\paragraph{Baselines.}  
In our study, we compare ComFusion with several state-of-the-art (SOTA) methods:
\begin{itemize}[leftmargin=*]
\item \textbf{DreamBooth}~\cite{ruiz2023dreambooth}: A SOTA approach that fully finetunes all layers of UNET and text-encoder. 
\item \textbf{Textual-Inversion (TI)}~\cite{gal2022image}: A SOTA approach that only  focuses solely on training word embeddings. 
\item \textbf{Custom-Diffusion (CD)}~\cite{kumari2023multi}: A concurrent work optimizing the cross-attention weights of the denoising model, along with a newly-added text token. Official hyperparameters are utilized.
\item \textbf{Extended Textual-Inversion (XTI)}~\cite{voynov2023p+}: Building on TI~\cite{gal2022image}, XTI inverts input images into a set of token embeddings, one per layer, demonstrating faster, more expressive, and precise results than TI. 
\item \textbf{ELITE}~\cite{wei2023elite}: This method trains a local and global map on large-scale datasets, enabling the instant generation of new images from a single user-provided image and corresponding mask.
\item \textbf{Break-A-Scene}~\cite{avrahami2023break}: This method employs a two-stage training strategy, optimizing token embedding, text-encoder, and UNET under the supervision of an object mask.
\end{itemize}
It's noteworthy that existing personalization methods, such as DreamBooth~\cite{ruiz2023dreambooth}, TI~\cite{gal2022image}, CD~\cite{kumari2023multi}, and XTI~\cite{voynov2023p+} typically require multiple images as input, in contrast to Break-A-Scene~\cite{avrahami2023break} and ELITE~\cite{wei2023elite}, which leverage a single image with a mask indicating the target concept.
In our setting, we use a single instance image without a mask to generate new images featuring the target concept in multiple specific scenes. 
For our experiments, unless stated otherwise, we employ the 30-step DDIM~\cite{song2020denoising} sampler with a scale of 7.5

\paragraph{Experimental Settings.} 
For the methods mentioned, including DreamBooth~\cite{ruiz2023dreambooth}, CD~\cite{kumari2023multi}, and our proposed ComFusion, all of which utilize class-aware prior images, we generate $200$ prior images to ensure a fair comparison. Besides, Break-A-Scene~\cite{avrahami2023break} relies on instance masks for its training process, while ELITE~\cite{wei2023elite} depends on instance masks during inference. Therefore, for these methods, we obtain the concept mask of the instance image using SAM~\cite{kirillov2023segment}.
All experiments are conducted on a single A100 GPU. For all pre-trained Stable Diffusion (SD) models, we use the 1.5 checkpoint~\cite{rombach2022high} for those baseline methods except for ELITE~\cite{wei2023elite} without training.
Here are the detailed settings for each method: 

\begin{itemize}[leftmargin=*]

\item\textbf{ComFusion}:  
ComFusion uses a pre-trained Stable Diffusion (SD) checkpoint 1.5~\cite{rombach2022high} and produce $200$ prior images, and finetunes text-encoder $\Gamma$ and denoising model $\epsilon_{\theta}$ architectured with $UNET$~\cite{ronneberger2015u} for 1200 steps, using a batch size of $1$ and learning rate $1 \times 10^{-5}$. 
During training, the hyper parameters $\lambda_C^S$ (\emph{resp.,} $\lambda_F^S$, $\lambda_F^I$,$\tau$) is set as $1$ (\emph{resp.,} $0.01$, $0.01$, $3$).

\item\textbf{DreamBooth~\cite{ruiz2023dreambooth}}: Similar to ComFusion, DreamBooth uses instance images and $200$ prior images to finetune the text-encoder and denoising model $UNET$ based on Stable Diffusion (SD) checkpoint 1.5~\cite{rombach2022high}. The total training steps are $1200$, set the batch size (\emph{resp.,} learning rate) as $1$ (\emph{resp.,} $1 \times 10^{-5}$) .

\item\textbf{TI~\cite{gal2022image}}: Based on Stable Diffusion (SD) checkpoint 1.5~\cite{rombach2022high}, TI leverages instance images to learn a token embedding with a batch size of $4$. The base learning rate was set to $0.005$ and the model is trained with 5, 000 optimization steps.

\item\textbf{CD~\cite{kumari2023multi}}: Following original setting in ~\cite{kumari2023multi}, CD loads a pretrained Stable Diffusion (SD) checkpoint 1.5~\cite{rombach2022high}. CD~\cite{kumari2023multi} learns a new token embedding and finetunes the $UNET$ parameters with 250 steps on the combination of instance image and prior images. The batch size is set as $8$ and learning rate is $8 \times 10^{-5}$. During training, training images are randomly resized for augmentation. 

\item\textbf{XTI~\cite{voynov2023p+}}: Following original setting in ~\cite{voynov2023p+},  XTI adopts a reduced learning rate of $0.005$ without scaling for optimization with batch size of $8$, the model is trained for $500$ steps to learn new token embeddings.

\item\textbf{ELITE~\cite{wei2023elite}}: ELITE is a pretrained model and can be instantly applied for generating new images with input of instance image and its mask.

\item\textbf{Break-A-Scene~\cite{avrahami2023break}}: Following original setting in ~\cite{avrahami2023break}, we load a pretrained Stable Diffusion (SD) checkpoint 1.5~\cite{rombach2022high}, the Break-A-Scene~\cite{avrahami2023break} adopt two-stage training strategy: in the first stage only the text embeddings is optimized with a high learning rate of $5 \times 10^{-4}$, while in the second stage, both the $UNET$ weights and the text encoder weights are optimized with a small learning rate of $2 \times 10^{-6}$. Both stages use Adam optimizer. Each stage is trained for $400$ steps.
\end{itemize}

\section{Evaluation Metrics} \label{sec:evaluation_metric_details}
To assess the fidelity of both instances and scenes in the generated images, we conduct both quantitative and qualitative evaluations.
Following DreamBooth~\cite{ruiz2023dreambooth}, we use DINO score~\cite{caron2021emerging}, and CLIP-I~\cite{radford2021learning} to evaluate instance fidelity, and use CLIP-T~\cite{radford2021learning} to evaluate the scene fidelity.
\begin{itemize}[leftmargin=*]
\item \textbf{CLIP-I}~\cite{ruiz2023dreambooth}: Measures the average pairwise cosine similarity between CLIP~\cite{radford2021learning} embeddings of generated and real images.
\item \textbf{DINO}~\cite{caron2021emerging}: Calculates the average pairwise cosine similarity using ViT-S/16 DINO~\cite{caron2021emerging} embeddings of generated and real images. Unlike supervised networks, DINO does not ignore differences within the same class but rather focuses on distinct features of a subject or image, thanks to its self-supervised training objective.
\item \textbf{CLIP-T}~\cite{ruiz2023dreambooth}: This metric evaluates the alignment between the textual prompts and the image ~\cite{radford2021learning} embeddings, thereby assessing the fidelity of the input scene as represented in the generated images.
\end{itemize}

\section{Datasets} \label{sec:dataset_introduction}
The dataset used in this paper are 25 concepts from DreamBooth Dataset and TI~\cite{gal2022image} dataset. 
The single concept image from two dataset are visualized in ~\cref{fig:datasets}.
The $15$ specific instance-scenes are: ``\texttt{[identifier] [class noun] Scene}'', the specific scenes including: ``\texttt{in the rain}'', ``\texttt{in the river}'', ``\texttt{in the sky}'', ``\texttt{in the room}'', ``\texttt{in the basket}'', ``\texttt{in the TV}'', ``\texttt{in the snow}'', ``\texttt{on the sofa}'',  ``\texttt{on the bed}'', ``\texttt{on the table}'', ``\texttt{on the stage}'', ``\texttt{on the top of mountain}'',  ``\texttt{on the playground}'',  ``\texttt{on the floor}'', ``\texttt{on the grass}''.

\begin{figure*}[htp]
\begin{center}
\includegraphics[width=1\linewidth]{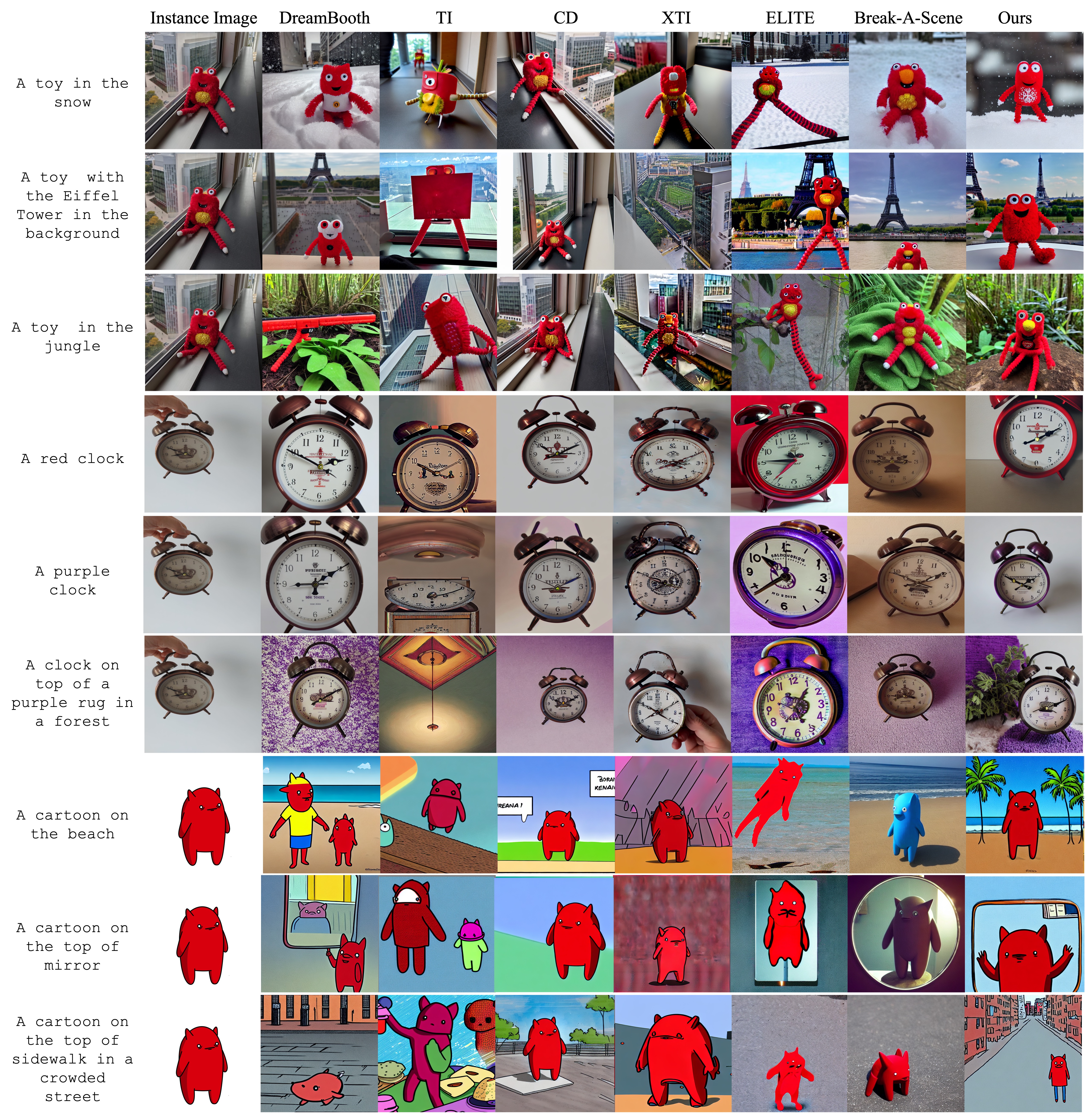}
\end{center}
\vspace{-16pt}
\caption{Images in unseen scenes generated by baseline methods and our proposed ComFusion trained from a single instance image.}
\vspace{-8pt}
\label{fig:beyond_specific} 
\end{figure*}

\begin{table}[!htp]
  \centering
  \caption{Generalization to unseen scenes. Quantitative metric comparison of instance fidelity (DINO, CLIP-I) and scene fidelity (CLIP-T).} 
  \vspace{-8pt}
   \resizebox{0.5\columnwidth}{!} {
  \begin{tabular}{l|ccc}
\toprule 
Methods&  DINO ($\uparrow$)& CLIP-I ($\uparrow$)&  CLIP-T ($\uparrow$)\\
\midrule 
Real Images&\textit{0.795} & \textit{0.859}&N/A \\
\midrule
DreamBooth~\cite{ruiz2023dreambooth}&0.607&0.735&0.214  \\
TI~\cite{gal2022image} &0.459&0.632&0.188 \\
CD~\cite{kumari2023multi}&0.611&0.725&0.202  \\
XTI~\cite{voynov2023p+}&0.431&0.602&0.185  \\
ELITE~\cite{wei2023elite}&0.415&0.607&0.241  \\
Break-A-Scene~\cite{avrahami2023break} &0.618&0.749&0.261  \\
Ours&\bf{0.621}&\bf{0.752}&\bf{0.297}  \\
\bottomrule
\end{tabular}
}
  \label{tab:beyond_specific}
\end{table}

\begin{figure*}[htp]
\begin{center}
\includegraphics[width=1\linewidth]{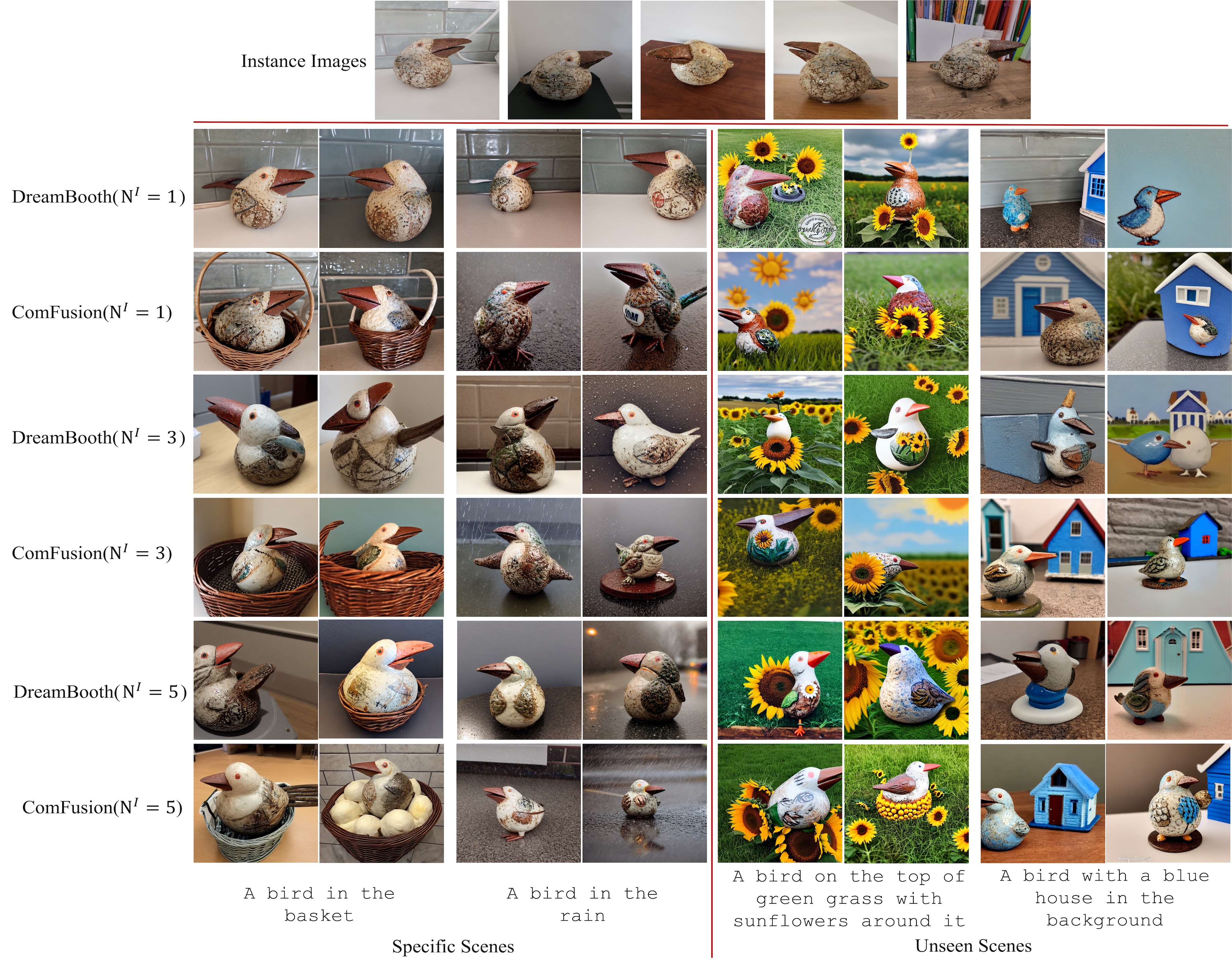}
\end{center}
\caption{Images generated by DreamBooth~\cite{ruiz2023dreambooth} and our proposed ComFusion in specific scenes (the left four column) and unseen scenes (the right four column). }
\label{fig:multi_instance} 
\end{figure*}
\begin{table}[!htp]
  \centering
  \caption{ComFusion trained on multiple instance images, and testing in specific scenes. Quantitative metric comparison of instance fidelity (DINO, CLIP-I) and scene fidelity (CLIP-T).} 
  \vspace{-8pt}
   \resizebox{0.5\columnwidth}{!} {
  \begin{tabular}{l|ccc}
\toprule 
Methods&  DINO ($\uparrow$)& CLIP-I ($\uparrow$)&  CLIP-T ($\uparrow$)\\
\midrule 
Real Images&\textit{0.795} & \textit{0.859}&N/A \\
\midrule
DreamBooth ($N^I$=1)&0.619&0.752& 0.229 \\
Ours ($N^I$=1)& \bf{0.658} & \bf{0.814} & \bf{0.321} \\
\hline
DreamBooth ($N^I$=3)&0.639&0.791&0.246\\
Ours ($N^I$=3)&\bf{0.669}&\bf{0.834}&\bf{0.332}\\
\hline
DreamBooth ($N^I$=5)&0.629&0.761&0.261\\
Ours ($N^I$=5)&\bf{0.661}&\bf{0.825}&\bf{0.348}\\
\bottomrule
\end{tabular}
}
  \label{tab:multiple_instance}
\end{table}

\begin{table}[!htp]
  \centering
  \caption{ComFusion trained on multiple instance images, and testing in unseen scenes. Quantitative metric comparison of instance fidelity (DINO, CLIP-I) and scene fidelity (CLIP-T).} 
  \vspace{-8pt}
   \resizebox{0.5\columnwidth}{!} {
  \begin{tabular}{l|ccc}
\toprule 
Methods&  DINO ($\uparrow$)& CLIP-I ($\uparrow$)&  CLIP-T ($\uparrow$)\\
\midrule 
Real Images&\textit{0.795} & \textit{0.859}&N/A \\
\midrule
DreamBooth ($N^I$=1)&0.607&0.735& 0.214 \\
Ours ($N^I$=1)& \bf{0.618} &  \bf{0.749} &  \bf{0.297} \\
\hline
DreamBooth ($N^I$=3)&0.613&0.752&0.219 \\
Ours ($N^I$=3)& \bf{0.640}&  \bf{0.767}&  \bf{0.301} \\
\hline
DreamBooth ($N^I$=5)&0.611&0.748&0.231 \\
Ours ($N^I$=5)& \bf{0.622}& \bf{0.753}& \bf{0.306} \\
\bottomrule
\end{tabular}
}
  \label{tab:multiple_instance_unseen}
\end{table}

\begin{figure*}[htp]
\begin{center}
\includegraphics[width=1.0\linewidth]{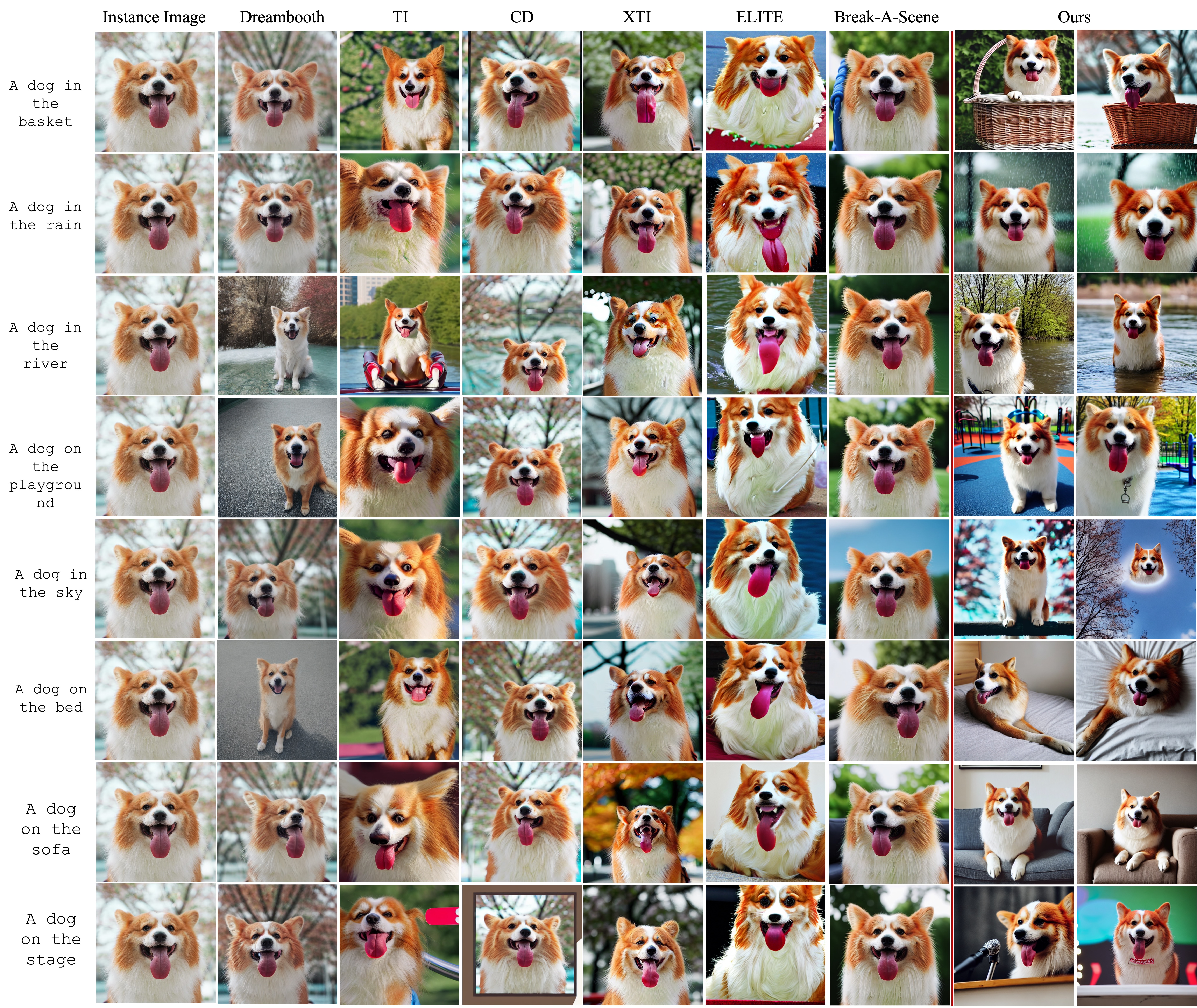}
\end{center}
\caption{Images generated by DreamBooth~\cite{ruiz2023dreambooth}, TI~\cite{gal2022image},CD~\cite{kumari2023multi},XTI~\cite{voynov2023p+},ELITE~\cite{wei2023elite}, Break-A-Scene~\cite{avrahami2023break}, and our proposed ComFusion in multiple specific scenes from a single instance image.}
\label{fig:supple_comparison_dog} 
\end{figure*}
\begin{figure*}[htp]
\begin{center}
\includegraphics[width=1.0\linewidth]{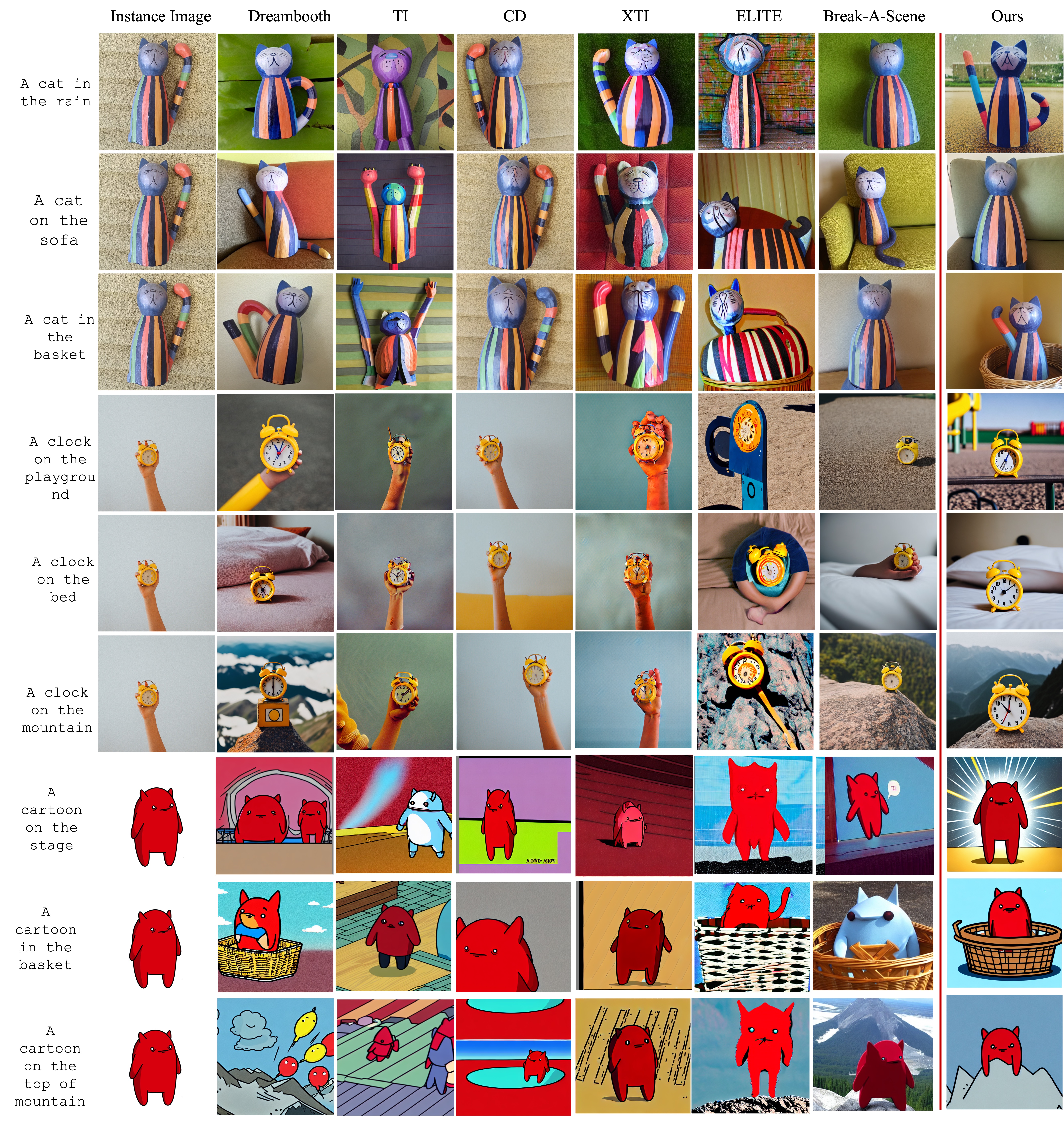}
\end{center}
\caption{Images generated by DreamBooth~\cite{ruiz2023dreambooth}, TI~\cite{gal2022image},CD~\cite{kumari2023multi},XTI~\cite{voynov2023p+},ELITE~\cite{wei2023elite}, Break-A-Scene~\cite{avrahami2023break}, and our proposed ComFusion in multiple specific scenes from a single instance image.}
\label{fig:supple_comparison} 
\end{figure*}
\begin{figure*}[htp]
\begin{center}
\includegraphics[width=1\linewidth]{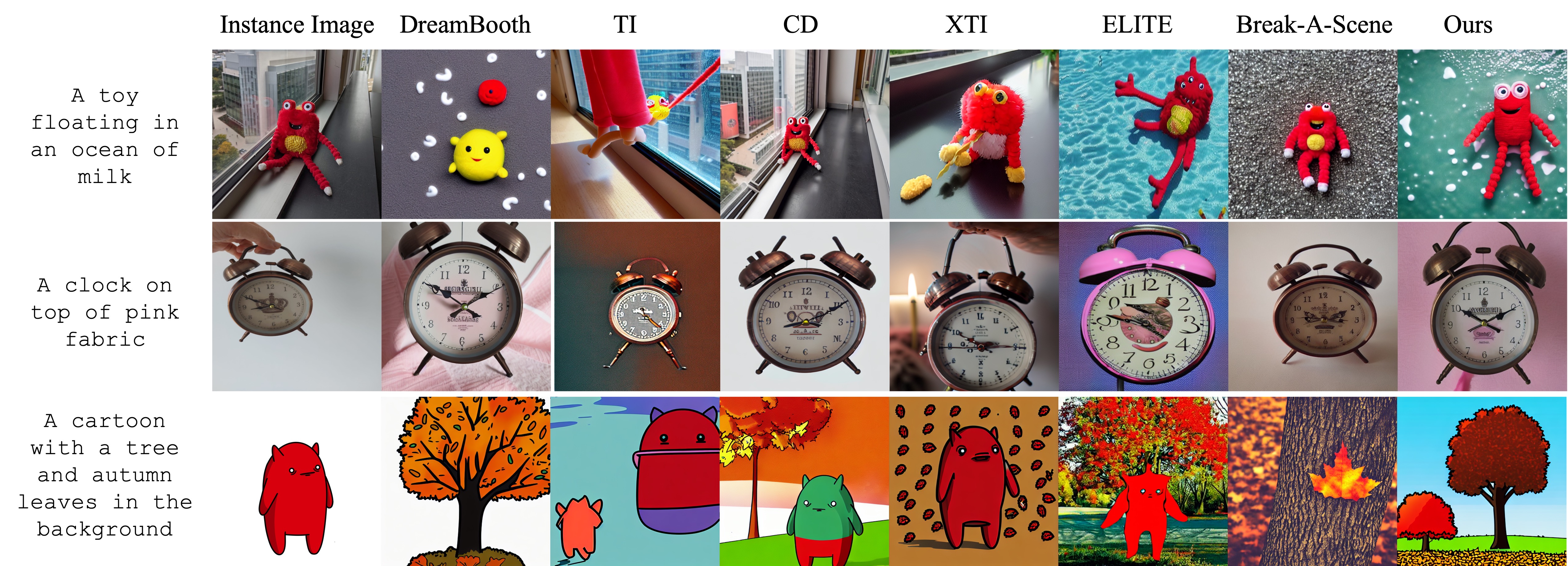}
\end{center}
\caption{Failure cases in unseen scenes.}
\label{fig:limitation} 
\end{figure*}

\section{Generalization to Unseen Scenes}\label{sec:unseen_scene}
To assess ComFusion's capability in generating images for unseen scenes, we follow  DreamBooth~\cite{ruiz2023dreambooth} by using $25$ diverse prompts, which include $20$ recontextualization prompts and $5$ property modification prompts. For each prompt, ComFusion and the baseline methods are employed to sample $10$ images. The instance fidelity and scene fidelity of these images are then evaluated using CLIP-I, CLIP-T, and DINO metrics.
Quantitative comparison results are reported in ~\cref{tab:beyond_specific}, while qualitative outcomes are illustrated in ~\cref{fig:beyond_specific}. The results from \cref{tab:beyond_specific} indicate that ComFusion achieves the highest scene fidelity scores, maintaining instance fidelity comparable to Break-A-Scene\cite{avrahami2023break}.
This performance can be attributed to the integration of class-scene prior images in the training process, which supplements the model with additional textual information. This enhancement aids the model in generalizing to unseen scenes and mitigates the risk of overfitting to the specific prompt structure “\texttt{a [identifier] [class noun]}”. However, the combination of instances with unseen scenes, which is not encountered during training, may result in a slightly lower instance fidelity score.

\section{Multiple Instance Images} \label{sec:multiple_instance}
In the main paper, we utilize a single instance image to train ComFusion, aiming to assess its few-shot learning ability. To further evaluate the impact of the number of instance images $\bm{x}^I$ on ComFusion's performance, we conduct additional experiments. In these tests, we maintain a constant number of class-scene images at $N = 200$ while varying the number of instance images $N^I$. Specifically, we explore scenarios where $N^I$ is set to either $3$ or $5$, allowing us to observe how changes in the number of instance images influence the effectiveness of our model in few-shot learning contexts.

\paragraph{Generalization on Specific Scenes.}
In accordance with the experimental setting described in Sec. 4.1 in main paper, we generated 10 images for each of the 25 subjects across each of the 15 scenes, resulting in a total of 3750 images for evaluation. 
Additionally, we calculated the CLIP-I, CLIP-T, and DINO metrics to assess both instance fidelity and scene fidelity, as detailed in ~\cref{tab:multiple_instance}.
From the table, it is evident that the proposed ComFusion model surpasses DreamBooth in performance when the number of instance images increases.
A notable trend observed is the enhancement in scene fidelity, as indicated by the CLIP-T score, with the increase in the number of instance images. However, this trend is not mirrored in the instance fidelity metrics (CLIP-I and DINO), where the scores for ``DreamBooth ($N^I=3$)'' (\emph{resp.,} ``ComFusion ($N^I=3$)'') are higher than for ``DreamBooth ($N^I=5$)''  (\emph{resp.,} ``ComFusion ($N^I=5$)''). 
We hypothesize that the use of multiple instance images can reduce overfitting to a specific instance image and introduce greater diversity to the target concept. 
This hypothesis is supported by the visual evidence in ~\cref{fig:multi_instance}, which shows a rich variety of bird poses and shapes when models are trained on either $3$ or $5$ instance images. 
This alleviation of overfitting, thanks to multiple instance images, also helps the pretrained model retain prior knowledge, thus achieving higher scene fidelity.

\paragraph{Generalization on Unseen Scenes.}
Expanding on the $25$ unseen scenes described in ~\ref{sec:unseen_scene}, we generated $10$ images for each of these scenes and assessed instance fidelity and scene fidelity using the CLIP-I, DINO, and CLIP-T metrics. 
A comparison between ~\cref{tab:multiple_instance} and ~\cref{tab:multiple_instance_unseen} reveals that the overall performance in unseen scenes is not as promising as in specific scenes. 
Analyzing ~\cref{fig:multi_instance} and ~\cref{tab:multiple_instance_unseen}, we observe a trend consistent with the findings in specific scenes. 
Specifically, ComFusion surpasses DreamBooth in performance when an equal number of instance images are used. 
When we increase the number of instance images from $1$ to $5$, there is a noticeable improvement in scene fidelity as evaluated by the CLIP-T metric. 
The best results for instance fidelity are achieved when the model is trained on $3$ instance images.

\section{More Visualization Comparison} \label{sec:visualization_supple}
In this section, we present more visualization comparisons as shown in ~\cref{fig:supple_comparison_dog} and ~\cref{fig:supple_comparison}.
Observing \cref{fig:supple_comparison_dog}, it's evident that images generated by ComFusion not only exhibit high instance accuracy but also align well with the input prompt in terms of background scene.
Break-A-Scene\cite{avrahami2023break} and CD~\cite{kumari2023multi} demonstrate strong instance fidelity, yet they lack diversity in the background and do not adequately respond to the input prompts.
DreamBooth\cite{ruiz2023dreambooth} tends to either replicate the instance image closely or generate scene-specific images with compromised instance fidelity. 
Both TI~\cite{gal2022image} and XTI~\cite{voynov2023p+} consistently struggle to accurately depict specific scenes described in the input prompts. 
ELITE~\cite{wei2023elite}, not being trained with instance images, falls short in instance fidelity compared to the other baseline methods.

\section{Limitations} \label{sec:limitation}
We visualize some failure cases in ~\cref{fig:limitation}, highlighting areas where both the baseline methods and ComFusion encounter challenges.
The first row of ~\cref{fig:limitation} demonstrates that both baseline methods and ComFusion struggle with understanding and rendering creative scenes, such as ``\texttt{in an ocean of milk}''. 
The second row shows that when it comes to descriptions of material properties (\emph{e.g.,} fabric), the methods exhibit limited capability in integrating instance concepts with such specific prompts. This suggests a gap in accurately representing detailed material textures and properties.
The third row highlights the challenge with long-term prompts that describe composite semantics, like a scene with a tree and autumn leaves in the background. Both baseline and proposed methods find it difficult to coherently integrate the target concept from the instance image with the background scene, often neglecting the target concept.
These limitations point to areas where further research and development could enhance the model's understanding and rendering capabilities, particularly in contexts involving creative, material, or composite semantic descriptions.

{\small
\bibliographystyle{ieeenat_fullname}
\bibliography{egbib}
}